%% file: wacv2021-A.tex
\documentclass[10pt,twocolumn,letterpaper]{article}

\usepackage{wacv}
\usepackage{times}
\usepackage{epsfig}
\usepackage{graphicx}
\usepackage{amsmath}
\usepackage{amssymb}
\usepackage{url}
\usepackage[normalem]{ulem}
\usepackage{booktabs} 
\usepackage{paralist} 
\usepackage{subfig}
\usepackage{tabularx}
\usepackage[table]{xcolor} 
\usepackage{paralist}
\usepackage{comment}
\usepackage{threeparttable}
\usepackage{algorithm}
\usepackage[noend]{algpseudocode}
\usepackage{mathrsfs}
\usepackage{color}

\newcommand{\para}[1]{\noindent\textbf{#1}~}

\newcommand{\miaojing}[1]{{#1}}

\def\wacvPaperID{1001} 

\wacvfinalcopy 

\ifwacvfinal
\fi

\ifwacvfinal
\usepackage[breaklinks=true,bookmarks=false]{hyperref}
\else
\usepackage[pagebackref=true,breaklinks=true,colorlinks,bookmarks=false]{hyperref}
\fi

\ifwacvfinal
\else
\fi

\begin{document}

\title{Fast Fourier Intrinsic Network}

\author{Yanlin Qian$^1$, Miaojing Shi$^2$\thanks{This work was partially done when Miaojing Shi was at INRIA Rennes and Yanlin Qian was an intern there. }, Joni-Kristian K{\"a}m{\"a}r{\"a}inen$^1$, Jiri Matas$^3$\\
$^1$Computing Sciences, Tampere University\\
$^2$Department of Informatics, King's College London\\
$^3$Center for Machine Perception, Czech Technical University in Prague\\
{\tt\small miaojing.shi@kcl.ac.uk}
}

\maketitle
\thispagestyle{empty}
\input{abstract.tex}

\input{fig_architecture.tex}
\input{introduction.tex}
\input{relatedwork.tex}

\input{methodology.tex}

\input{experiments.tex}

\input{conclusion.tex}

\clearpage

{\small

\bibliographystyle{ieee_fullname}
\bibliography{intrinsicimage}
}

\end{document}

%% file: abstract.tex
\begin{abstract}
We address the problem of decomposing an image into albedo and shading.
We propose the Fast Fourier Intrinsic Network,  FFI-Net in short, that
operates in the spectral domain, splitting the input into several spectral bands.
Weights in FFI-Net are optimized in the spectral domain, allowing faster
convergence to a lower error. FFI-Net is lightweight and does not need auxiliary networks for training.
The network is trained end-to-end with a novel spectral loss which measures the global distance between the network prediction and corresponding ground truth.
FFI-Net achieves state-of-the-art performance on MPI-Sintel, MIT Intrinsic, and IIW datasets.
\end{abstract}

%% file: fig_architecture.tex
\begin{figure*}[h]
\begin{center}
\includegraphics[width=0.9\linewidth]{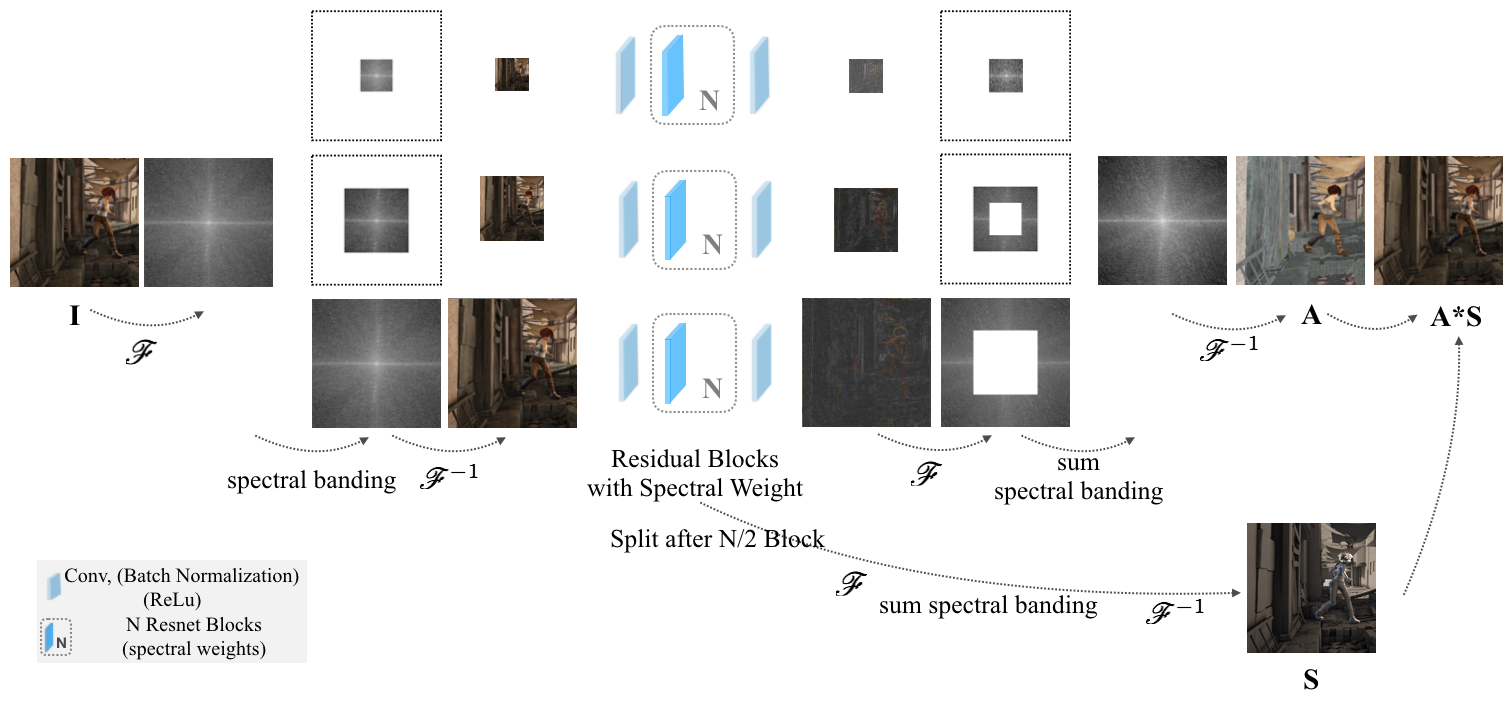}
\vspace{-5mm}
\caption{
Overview of FFI-Net. Different from other deep intrinsic image decomposition architectures, FFI-Net offers several new modules:
(a) $\mathscr{F}$ and $\mathscr{F}^{-1}$ layers that convert between spatial and spectral domains using
the Fast Fourier Transform (FFT) and its inverse (Sec. \ref{subsec:fft}) ; (b) \textit{spectral residual blocks} where residual block weights are parametrized in the spectral domain (Sec. \ref{subsec:ournetwork});
(c)
\miaojing{\textit{spectral banding} that splits a spectral map into $M$ (e.g. M = 3) overlapped (low-pass) spectral bands at the beginning of the network and associates $M$ non-overlapped spectral bands of the image in both spatial and spectral domain as ground truth by the end of the network.}
(d) \textit{spectral loss} which measures global error between the prediction and ground truth in the spectral domain (Sec. \ref{subsection:spectralloss}).}
\label{fig:architecture}
\end{center}
\vspace{-5mm}
\end{figure*}

%% file: introduction.tex
\section{Introduction}
Intrinsic image decomposition refers to the problem of separating an image $\mathcal{I}$ into its
\emph{albedo} and \emph{shading}~\cite{barrow1978recovering}, where \emph{albedo} models the diffuse reflection of the scene materials and 
\emph{shading} represents the lighting and scene geometry.
We adopt the common assumption that the problem can be simplified to the per-pixel image formation
$\mathcal{I}=\mathcal{A}\cdot\mathcal{S}$ where $\mathcal A$ and $\mathcal S$ are the
albedo and shading, respectively. 
Factorization of $\mathcal{I}$ into $\mathcal{A}$ and $\mathcal{S}$ is ill-posed.
Early work tackled the factorization by applying constraints and priors which connect the properties of
materials and lighting in the physical world with the gradient and color changes in images~\cite{land1971lightness,horn1974determining}; however,
even the sophisticated physical modelling can
not fully cover the  complex optical interactions among materials,
light sources, and object/scene geometry. 
Intrinsic image decomposition is an open and challenging problem {\em per~se}, but it can also benefit many other vision problems, \textit{e.g.}, {semantic segmentation}~\cite{baslamisli2018joint}, {depth estimation}~\cite{kim2016unified}, and {color constancy}~\cite{barron2015shape}.


Traditional approaches for intrinsic image decomposition are surveyed in~\cite{barron2015shape}, while recent state-of-the-art performance has been achieved with image-to-image translation networks as in~\cite{narihira2015direct,fan2018revisiting,cheng2018intrinsic,lettry2018darn,li2018learning,liu2020cvpr}. The mean squared error (MSE) loss is typically employed in the network training to enforce pixel-wise similarity between the predicted and ground truth albedo and shading~\cite{fan2018revisiting,lettry2018darn,cheng2018intrinsic}.
The MSE loss does not consider correlations between image pixels.  
Extra loss terms such as loss for the learnable domain filter~\cite{fan2018revisiting}, perceptual
loss~\cite{cheng2018intrinsic} and generative adversarial loss \cite{lettry2018darn}, are often added to enforce global similarities between predictions and ground truth. 
Proposed models also exploit a multi-path structure in their networks to separate the prediction into different levels of details~\cite{cheng2018intrinsic} or contexts~\cite{li2018learning,fan2018revisiting}.

In this work, we propose a \textbf{F}ast \textbf{F}ourier \textbf{I}ntrinsic \textbf{Net}work (FFI-Net)  for intrinsic image decomposition. 
The network design focuses on the multi-path architecture and loss function, as were intensively exploited in recent works~\cite{li2018learning,fan2018revisiting,lettry2018darn,cheng2018intrinsic}, but differs substantially from the prior arts since no published intrinsic decomposition method operates in the spectral domain.
The intuition is based on  the observation that published approaches
~\cite{fan2018revisiting,cheng2018intrinsic} often separate the predictions of structure and details into different scales, implicitly performing 
operations that are indeed best understood in the spectral domain. 
Image details, \eg fine edges, are naturally reflected in the high-frequency part;
on the other hand, image structures such as flattened color patches are usually reflected in the low-frequency part. 
Despite the frequency interpretation behind the previous works, their models were optimized in the spatial domain. 
Thus if the intrinsic image decomposition task is solved in the spectral domain, we might expect, 
and this is confirmed by our experiments, the optimization of the network to be more efficient and effective, as in the classification task ~\cite{rippel2015spectral}. 
Furthermore, since each pixel in the spectral domain is a complex-weighted sum over the image space,  a pixel-wise loss on the spectral map thus can straightforwardly enforce a global similarity between the prediction and ground truth in the spatial domain.     

\para{Contributions.}
We propose a novel network architecture for intrinsic image decomposition where the following spectral operations and structures replace their
spatial domain counterparts used in the prior works:
\begin{compactitem}
%
\item \textit{Spectral banding} splits the input image into several frequency bands, therefore allowing multi-path training and inference (see Fig.~\ref{fig:architecture}).

\item \textit{Spectral loss} provides a global error measure on the similarity between the predicted $\mathcal{A}$/$\mathcal{S}$ and the corresponding ground truth, effectively augmenting the pixel-wise loss.

\item \textit{Spectral residual block} encodes original residual block with  spectral weights, thus obtaining faster convergence to a lower error.

\end{compactitem}

To the best of our knowledge, FFI-Net is the first to address intrinsic decomposition in the Fourier domain.
Our experiments with three intrinsic image datasets (MPI-Sintel, MIL Intrinsic and IIW) demonstrate state-of-the-art accuracy for FFI-Net. 

%% file: relatedwork.tex
\section{Related Work}\label{sec:relatedwork}
Our main objective is to develop a deep neural network architecture
for intrinsic image decomposition and therefore related works
will focus on the prior deep 
architectures~\cite{narihira2015direct,kim2016unified, fan2018revisiting,lettry2018darn,cheng2018intrinsic,baslamisli2018cnn}. Traditional approaches have
been surveyed in the recent work by Barron and Malik~\cite{barron2015shape}.

Nahiriha~\etal~\cite{narihira2015direct} were the first to propose
a deep architecture to estimate albedo and
shading using a pixel-wise MSE loss. Kim~\etal~\cite{kim2016unified} later
added a conditional random field (CRF) after the network to jointly learn albedo, shading and depth.
Lettry~\etal~\cite{lettry2018darn} further proposed to add an adversarial module. To better model the flattening effects in natural reflectance images, Fan \etal \cite{fan2018revisiting} proposed a 1D recursive domain filter and 
separate loss layers customized differently for dense
(\eg MPI-Sintel~\cite{butler2012naturalistic}) and sparse (\eg IIW~\cite{bell2014intrinsic})
datasets; Cheng~\etal~\cite{cheng2018intrinsic} proposed
to use a cross bilateral filtering loss instead of the MSE loss to ensure the smoothness of albedo and shading output. 
Recently, Baslamisli~\textit{et al.}~\cite{baslamisli2018cnn} adopted a physics-based image formation model inspired by traditional approaches.

The above works assume the Lambertian reflectance model and 
are fully-supervised with ground truth albedo and shading. There are also many works assuming a non-Lambertian model~\cite{shi2017learning}, adopting the weakly-supervised/unsupervised
setting~\cite{janner2017self,li2018learning,ma2018single,liu2020cvpr}, or including more synthetic data for training~\cite{cgintrinsic,zhou2019glosh}. For instance, Shi \etal
\cite{shi2017learning} considered the non-Lambertian model in their algorithm design and train their network with rendered 3D models from ShapeNet~\cite{chang2015shapenet}. 
Li \etal \cite{li2018learning} learned intrinsic images from image sequences over time, where the object reflectance remains the same with varying illumination.
Liu \etal \cite{liu2020cvpr} proposed a novel unsupervised intrinsic image decomposition
framework which directly learns the latent feature of reflectance and shading from unsupervised and
uncorrelated data. Zhou \etal \cite{zhou2019glosh} proposed a global-local spherical harmonics lighting model for intrinsic decomposition, where they 
apply synthetic data for model pre-training and fine-tune it with real data in a self-supervised way.

The FFI-Net is inspired by several recent works while the most important being
~\cite{cheng2018intrinsic, rippel2015spectral}. FFI-Net differs from Cheng \etal~\cite{cheng2018intrinsic} rather substantially in the following ways:
(i) their hierarchical decomposition is done in the spatial domain with fixed-sized Gaussian kernels, while the decomposition in FFI-Net is
made in the spectral domain;
(ii) they use VGG perceptual loss to enforce global similarity between the intrinsic prediction and ground truth while FFI-Net uses the basic
MSE in the spectral domain;
(iii) their network optimization is done in the spatial domain while ours in the spectral domain. With respect to the comparison to~\cite{rippel2015spectral}, FFI-Net adopts the spectral encoding of CNN weights from it but novelly introduces several new elements \ie spectral banding, spectral loss, spectral residual block, for intrinsic image decomposition. 

%% file: methodology.tex
\section{Fast Fourier Intrinsic Network}\label{sec:method}
Before presenting the building blocks of FFI-Net, we review the properties of the Discrete Fourier Transform.
\subsection{Preliminaries}
\label{subsec:fft}


%
\para{The Discrete Fourier Transform (DFT).} 
For a discrete 2D image $I(h,w) = I \in \mathbb{R}^{H \times W}$ where
$h=0,\ldots,H-1$ and $w=0,\ldots, W-1$, the
Fourier domain representation is computed by 
%
\begin{align}
& \mathscr{F}(I)_{nm} = \frac{1}{\sqrt{HW}}\sum_{h=0}^{H-1} \sum_{w=0}^{W-1} I(h,w) e^{-2\pi i(\frac{nh}{H} + \frac{mw}{W})}\qquad \notag \\
& \forall n\in\{0,\ldots, H-1\}, \forall m\in\{0,\ldots, W-1\}\;.  
\end{align}
%
The DFT is linear and unitary and $I$ can be restored using the inverse DFT given by ${\mathscr{F}^{-1}(I)=\mathscr{F}(I)^*}$, which is the conjugate of the transform itself. 

\vspace{\medskipamount}
\para{Properties of DFT for CNN.}\\ 
\noindent\emph{Convolution theorem:}
Convolution in the spatial domain is equal to the element-wise product in the spectral domain: 
\begin{align}
\mathscr{F}(x*f) = \mathscr{F}(x)\odot\mathscr{F}(f),
\end{align}
where we denote by $x$ and $f$ two spatial signals (\eg images), $*$ and $\odot$ the operators of the convolution and the Hadamard product, respectively. 
DFT can be implemented using the Fast Fourier Transform (FFT), 
which reduces the complexity of DFT from $\mathcal{O}((HW)^{2})$ to $\mathcal{O}(HW \log (HW))$.

\medskip
\noindent\emph{DFT differentiation:} For CNN backpropagation algorithms, the gradients must be
propagated through Fourier transform layers. Suppose a DFT layer $\mathscr{F}$ maps a real-valued input $x \in \mathbb{R}^{H \times W}$ to $z \in \mathbb{C}^{H \times W}$, and $\frac{\partial{\mathcal L}}{\partial{z}}$ is the gradient of loss $\mathcal L$ w.r.t $z$; Since both $\mathscr{F}$ and $\mathscr{F}^{-1}$ are unitary, the back-propagation through $\mathscr{F}$ and $\mathscr{F}^{-1}$ can be written as 
\begin{align}
\frac{\partial \mathcal L}{\partial{x}} = \mathscr{F}^{-1}\left(\frac{\partial \mathcal L}{\partial{z}}\right)\;,
~~~
\frac{\partial \mathcal L}{\partial{z}} = \mathscr{F}\left(\frac{\partial \mathcal L}{\partial{x}}\right)\;.
\label{eq:gradient}
\end{align}

\subsection{The Network Architecture}
\label{subsec:ournetwork} 
Our network architecture is built upon the preliminary discussion about DFT and extended to the intrinsic image decomposition task with several new components (see Fig.~\ref{fig:architecture}). The input image $I$ is first transformed into the spectral representation $\mathscr{F}({I})$. Spectral banding (Sec.~\ref{subsec:spectralbanding}) is applied afterwards to crop the spectral feature map of $\mathscr{F}({I})$ into $M$ overlapped (low-pass) sub-representations. DC component is centered in the spectral feature map in Fig.~\ref{fig:architecture}. 
The $M$ spectral bands are transformed back to the spatial domain via $\mathscr{F}^{-1}$ as input to feed into $N$ consecutive spectral residual blocks, respectively (Fig.~\ref{fig:spectralresidualblock} and Sec.~\ref{subsec:residual}). \miaojing{Spectral residual blocks parametrize the conv weights in the spectral domain for optimization yet their input and output are in the spatial domain. The outputs of the $M$ subbranches are transformed to the spectral domain, ground truth in two domains are associated both before and after the transformation. The $M$ ground truth are generated via the spectral banding that is different from previous: it produces a series of non-overlapped bands similar to the Laplacian pyramid (Sec.~\ref{subsec:spectralbanding}). The $M$ spectral maps predicted in the network are summed up to create a complete spectral map of the image and further converted back to the spatial domain; again, ground truth are associated both in the spectral and spatial domain on the complete image.  
The loss function between the ground truth and network predictions are introduced in the spatial and spectral domain, respectively (Sec.~\ref{subsection:spectralloss} and~\ref{subsec:loss}): the spatial loss employs the commonly used pixel-wise MSE in the original images and their gradient maps~\cite{fan2018revisiting,lettry2018darn} (Sec.~\ref{subsec:loss}); the spectral loss also utilizes pixel-wise MSE loss but on the spectral maps, which provides an additional global constraint on the intrinsic prediction (Sec.~\ref{subsection:spectralloss}).}
Notice that, to infer both the albedo and shading simultaneously, after the $\frac{N}{2}$-th block of each subbranch we split it into two streams for albedo and shading, respectively.  

Below we present the spectral residual block, spectral banding, and spectral loss.



\subsection{Spectral Residual Block} \label{subsec:residual}
\begin{figure}
\begin{center}
\includegraphics[width=1.0\linewidth]{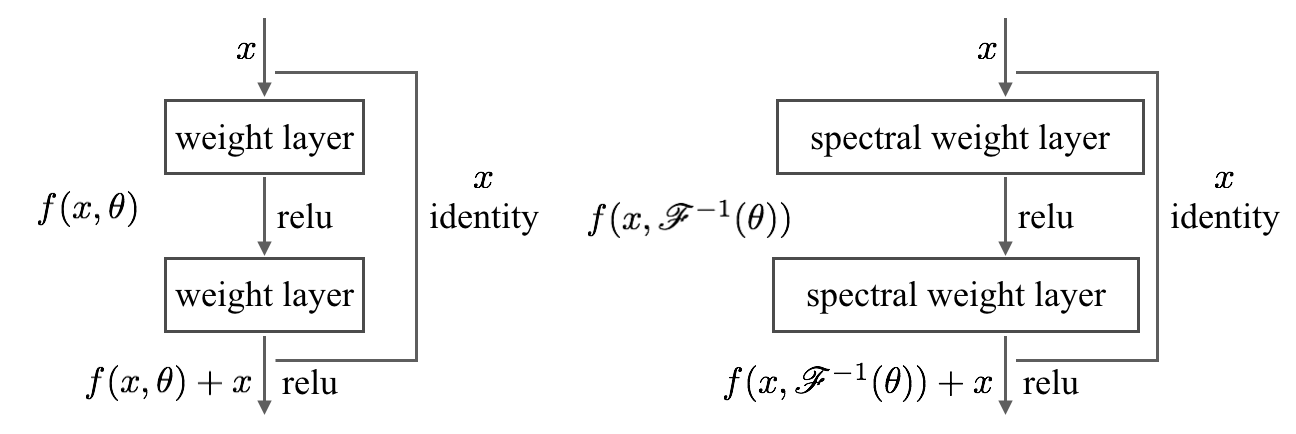}
\caption{ 
Illustration of the original (left) residual block and proposed spectral residual block (right). $\theta$ denotes the kernel weights of convolutional filters. 
\label{fig:spectralresidualblock}
}
\vspace{-5mm}
\end{center}
\end{figure}

Residual blocks~\cite{he2015deep} are widely applied for intrinsic image decomposition~\cite{fan2018revisiting,lettry2018darn}. Inspired by~\cite{rippel2015spectral}, we expect that spectral encoding of residual blocks can help solve the intrinsic decomposition task.
As shown in Fig.~\ref{fig:spectralresidualblock}, given a two-conv layer residual block, \miaojing{its input and output are in the spatial domain but the conv weights are parametrized in the spectral domain. These weights need to be converted back to the spatial domain when convolving with the input. This reparametrization does
not alter the underlying model, it provides an appropriate domain for parameter optimization, as the frequency basis captures typical filter structure well.
As suggested in~\cite{rippel2015spectral}, spectral encoding of the network results in more compact convolutional filters with more meaningful axis-aligned directions for optimization. 
In the experiment we demonstrate that residual blocks with spectral weights boost the performance.}

\subsection{Spectral Banding}
\label{subsec:spectralbanding}

Inspired by the multi-path structure in~\cite{cheng2018intrinsic}, which conducts scale space separation of the albedo/shading image, we instead perform a spectral domain separation, which provides more fine-grained control over frequency. 
The spectral banding is formulated as:
\begin{align}
\label{eq:spectralbanding}
    z_{nm}=
    \begin{cases}
      \mathscr{F}(x)_{nm}, & \text{if}\ n \in [N_l,N_u], m \in [M_l,M_u] \\
      0, & \text{otherwise} 
    \end{cases}
\end{align}

In a spectral map, values between $N_l$ and $N_u$ in $x$-axis, $M_l$ and $M_u$ in $y$-axis are kept. If $N_l$ and $M_l$ are always zeros, \textit{spectral banding} works like a smoothing filter. 
\miaojing{We apply this special case of spectral banding ($N_l$ and $M_l$ are zeros while $N_u$ and $M_u$ are increased over bands) at the beginning of the network to split the input into overlapped bands (see Fig.~\ref{fig:architecture} before the spectral residual blocks).} If $N_l$ and $M_l$ are non-zeros, \miaojing{we can separate the spectral map into multiple bands where the upper bounds ($N_u$, $M_u$) of a current band are taken as the lower bounds ($N_l$, $M_l$) of its adjacent band with higher frequency.} When we apply the inverse FFT to these spectral bands, they look like image Laplacian pyramid with different levels of details~\cite{burt1983tc,liu2020acmmm} (see Fig.~\ref{fig:architecture} after the spectral residual blocks). In the network, we generate $M$ of these non-overlapped spectral bands from the ground truth to associate at the end of each subbranch. In this way, the subbranch output can be summed together to recover the final prediction (we call it \textit{summing spectral bands} in Fig.~\ref{fig:architecture}). 

In practice, we can provide a specific range of $nm$ for $z_{nm}=0$ in Eq.~\ref{eq:spectralbanding}, where the spatial transform of a spectral band can be of adjustable size related to the predefined range. This makes spectral banding flexible with any input and output size in a fully-convolutional neural network. Notice the {level-of-detail} exploited in~\cite{cheng2018intrinsic} is obtained by applying the fix-sized Gaussian kernel in the spatial domain, while spectral banding retrieves the level-of-detail in the spectral domain explicitly and more effectively; \miaojing{Most importantly, our spectral banding enables the effective usage of another new component, spectral loss, as below.   }



\subsection{Spectral Loss}
\label{subsection:spectralloss}
Loss functions are defined in both the spectral and spatial domains, here we first detail the spectral loss: since each pixel in the spectral domain is a weighted sum of the whole image in the spatial domain, a pixel-wise loss defined on the spectral representation spontaneously enforces a global similarity between the spatial prediction and the ground truth. Inspired by the global loss applied  in~\cite{lettry2018darn,fan2018revisiting,cheng2018intrinsic}, we propose a spectral loss defined in the spectral domain:  
\begin{align}\label{eq:spectralloss}
\mathcal{L}_s(x,\hat{x}) = \|\mathcal{M}(\mathscr{F}(x))-\mathcal{M}(\mathscr{F}(\hat{x}))\|_2^2 
\end{align}
where $\hat{x}$ and $x$ denote the predicted and ground truth albedo/shading in the spatial domain, respectively. 
$\mathcal M$ is defined as an operator to take the magnitude of a complex value, $\mathcal{M}(\mathscr{F}(x)) = |\mathscr{F}(x)|$
We only use the magnitude part as it contains information of the image geometric structure in the spatial domain while discarding the phase part as ``invisible" \miaojing{(see ~\cite{R.fisher,dip_book}). This also empirically gives us benefits whilst minimizing the error over the phase part does not. Notice if we minimize the squared error for the real and imaginary components, since $\mathscr{F}$ is a linear transform, it will be identical to minimizing the squared error in the spatial domain, as specified below. }

\miaojing{In FFI-Net, spectral losses are directly applied pixel-wise on the $M$ spectral outputs of the network as well as on  their summation.  }


\subsection{Network Training}\label{subsec:loss}

Beside the \textit{spectral loss}, we also adopt representative local MSE loss in the spatial domain to let the predication to be both pixel-wise close and visually similar to the ground truth. Following~\cite{lettry2018darn,fan2018revisiting, narihira2015direct}, we choose the $\ell_2$ loss defined on every pixel value and its gradient:
\begin{align}
\mathcal{L}_p(x,\hat{x}) = \|x- \hat{x}\|_2^2 ,~~~ \mathcal{L}_g(x,\hat{x}) = \|\nabla(x)- \nabla\hat{x}\|_2^2.
\end{align}
Each of the $M$ subbranches in the network is trained with loss $\mathcal{L}_{ss}$ from both spectral and spatial domains:
\begin{align}
\mathcal{L}_{ss}(x,\hat{x}) = \lambda_{1}\mathcal{L}_p +  \lambda_{2}\mathcal{L}_g + \lambda_{3}\mathcal{L}_s. 
\end{align}
We empirically choose $\lambda_1=0.2,\lambda_2=0.35,\lambda_3=10^{-6}$ for all our experiments. \miaojing{$\mathcal{L}_s$ is of larger magnitude than $\mathcal{L}_g$ and $\mathcal{L}_p$ thus with a smaller weight $\lambda_3$.} $\mathcal{L}_{ss}$ is applied to all the $M$ subbranches as well as their final summation in the network, so we have the overall loss $\mathcal{L_T}$ for target albedo/shading:
\begin{align}
\mathcal{L_T} = \sum_{b=1}^{M}\mathcal{L}_{ss}^b(x_b,\hat{x}_b) + \mathcal{L}_{ss}(x,\hat{x}),
\end{align}
where we denote by $x_b$ and $\hat{x}_b$ the ground truth and prediction corresponding to the $b$-th spectral band of the original albedo/shading, respectively. 

Referring to Fig.~\ref{fig:architecture}, 
$\mathcal{L_T}$ is computed on both albedo and shading predictions ($\mathcal{L_A}$ and $\mathcal{L_S}$), so we have the final loss
\begin{align}
\mathcal{L} = \mathcal{L_A}  + \mathcal{L_S}  + \mathcal{L_{AS}}
\end{align}
$\mathcal{L_{AS}} = \|\mathcal{A}\cdot\mathcal{S} - \hat{\mathcal A}\cdot\hat{\mathcal S}\|_2^2 $ is added to address the commonly used constraint $\mathcal{I}=\mathcal{A}\cdot\mathcal{S}$ \cite{cheng2018intrinsic}.


%% file: experiments.tex
\section{Experiments}\label{sec:expresults}

\para{Datasets.}
We conduct experiments on three popular datasets: {\em MPI-Sintel~\cite{butler2012naturalistic}}, {\em MIT Intrinsic~\cite{grosse2009ground}} and {\em IIW}~\cite{bell2014intrinsic}.
MPI-Sintel~\cite{butler2012naturalistic} consists of $18$ animated video clips of multiple scenes, each containing $40$ or $50$ frames. We experiment
with two different data splits:
\textit{image split}~\cite{narihira2015direct,fan2018revisiting,cheng2018intrinsic} and
\textit{scene split}~\cite{narihira2015direct,cheng2018intrinsic}.
 MIT-Intrinsic~\cite{grosse2009ground} contains $20$ real objects in
$11$ different lighting configurations. The train/test split follows~\cite{barron2015shape}. IIW~\cite{bell2014intrinsic} has $5,230$ real images of mostly indoor scenes. Pairwise relative reflectance judgments are provided by humans as the ground truth albedo; the train/test split follows~\cite{fan2018revisiting}.  

\vspace{\medskipamount}

\para{Training Details.}
For all datasets, we fix the number of spectral bands to
$M\!=\!5$ and use $N\!=\!10$ spectral residual blocks
for each band. \miaojing{Spectral residual blocks have the same configuration as the ordinary residual block~\cite{he2015deep} in terms of filter size, batch normalization, relu \etc. Training is conducted using the Adam optimizer with network weights randomly initialized.}
\textrm{(i)} For MPI Sintel, we randomly crop $384\times 384$ patch from each image. 
The upper and lower bounds (Eq.\ref{eq:spectralbanding}) for each spectral band are $(384/4,384/2)$, $(384/8,384/4)$, $(384/16,384/8)$, $(384/32,384/16)$ and $(0,384/32)$, where $N_l = M_l$ and $N_u = M_u$ for square crops. We set the batch size to $4$ and the learning rate to \textrm{4e-4}. 
\textrm{(ii)} For MIT Intrinsic with smaller images, we randomly crop $192\times192$ patches and flip them. The upper and lower bounds for each band are $(192/4,192/2)$, $(192/8,192/4)$, $(192/16,192/8)$, $(192/32,192/16)$ and $(0,192/32)$, respectively. The batch size is $8$ and learning rate is \textrm{8e-3}. 
\textrm{(iii)} For IIW, we resize each image to $300\times 300$ as in~\cite{fan2018revisiting}; the batch size is set to $1$ and the learning rate is \textrm{1e-2}.
All networks are trained till convergences in our experiments.  


\vspace{\medskipamount}

\para{Evaluation Protocol.} 
 We report results using the four common metrics: \textrm{(i)} \textit{si-MSE}: scale-invariant mean squared error (MSE);  \textrm{(ii)} \textit{si-LMSE}: scale-invariant local MSE, which uses sliding windows in si-MSE; the window size $w$ is usually 10\% of the larger dimension of the image while the sliding step is $0.5w$; \textrm{(iii)} \textit{DSSIM}: dissimilarity structural similarity index measure; it is made to correlate with the subjective image quality assessments~\cite{wang2004image}. For MIT, we also report the MSE and LMSE numbers which can be seen as variants of si-MSE and si-LMSE~\cite{grosse2009ground}. For IIW with sparse ground truth, we report the weighted human disagreement rate (WHDR)~\cite{bell2014intrinsic,fan2018revisiting}. 
 
 We compare our method with a number of prior arts; these approaches are trained in various ways including fully-/weakly-supervised learning~\cite{fan2018revisiting,cheng2018intrinsic,cgintrinsic,lettry2018darn,ma2018single,zhou2019glosh}, unsupervised learning~\cite{liu2020cvpr,grosse09intrinsic,bell2014intrinsic,bi20151}, as well as learning from auxiliary datasets~\cite{narihira2015direct,shi2017learning,cgintrinsic,zhou2019glosh} (see Sec.~\ref{sec:relatedwork}). We do not elaborate every detail with every method but mainly focus our comparison with those representative fully-supervised methods, \eg ~\cite{fan2018revisiting,cheng2018intrinsic,cgintrinsic,lettry2018darn}.  

\input{tab_mpi_scenesplit.tex}
\input{fig-mpi-scenesplit.tex}

\subsection{Results on MPI-Sintel}
Results for FFI-Net and various competing works
are shown in Tab.~\ref{tab:sintel_scene_split} (scene-split) and Tab.~\ref{tab:sintel_image_split} (image-split). 

The scene-split setting is more challenging as it requires generalization over unseen scenes. Tab.~\ref{tab:sintel_scene_split} shows that FFI-Net performs the best on most metrics. In particular, it achieves the best
average si-MSE, si-LMSE, and DSSIM of albedo and shading: 1.18, 0.73, and 8.77, respectively.  

FFI-Net achieves the best performance also for the image-split setting reported in Tab.~\ref{tab:sintel_image_split}, but with smaller margin.
We suggest a possible reason as the overfitting of deep networks on the 
shared contents between training and test images in image-split. This is illustrated
by the recent results in~\cite{kim2016unified,fan2018revisiting,cheng2018intrinsic} from which the error numbers are indeed close to $0$ (see Tab.~\ref{tab:sintel_image_split}, they are multiplied by 100).

LapPyrNet~\cite{cheng2018intrinsic} performs in general the second best. It employs several loss functions such as joint bilateral filtering loss, VGG perceptual loss, \etc. 
Notwithstanding, our results show that the proposed simple spectral loss (Tab.~\ref{tab:sintel_scene_split}, only $\mathcal L_{\text{spec}}$) is as effective as these multiple losses .

Qualitative illustrations are
presented in Fig.~\ref{figure:mpi_scenesplit} and Fig.~\ref{figure:mpi_imagesplit} where in particular
the low-frequency part of the sky and high frequency details
of the person's face illustrate the improved prediction thanks to the spectral structure.

\input{tab_mpi_imagesplit.tex}
\input{fig-mpi-imagesplit.tex}
%
%
\subsection{Results on MIT Intrinsic}
\label{subsec:mit}


For MIT Intrinsic, \cite{lettry2018darn,fan2018revisiting,cheng2018intrinsic} report si-MSE and si-LMSE and while \cite{cgintrinsic, baslamisli2018cnn, li2018learning} report MSE, LMSE and DSSIM. We include both in Tab.~\ref{tab:mit} and~\ref{tab:mit_b} for comprehensive comparison.

With si-MSE and si-LMSE (Tab.~\ref{tab:mit}), FFI-Net outperforms the competing
deep learning methods. The traditional method SIRFS~\cite{barron2015shape} has 
slightly better si-MSE for albedo and total si-LMSE. It uses a prior that is particularly suitable for the MIT Intrinsic images; we notice its clear inferiority over others on MPI-Sintel in Tab.~\ref{tab:sintel_image_split}. 
MIT is a rather small dataset and is therefore
 often trained with additional datasets~\cite{shi2017learning,narihira2015direct} while FFI-Net only trains on MIT. 





\input{tab_mit_A.tex}
\input{tab_mit_B.tex}

\input{fig-mit.tex}

We also provide results in MSE, LMSE, and DSSIM in Tab.~\ref{tab:mit_b}.
FFI-Net performs in general better than latest works \eg~\cite{ma2018single,cgintrinsic,liu2020cvpr}. 
Notice that it appears that
the LMSE in \cite{ma2018single} is computed using a different window size, thus its two numbers (italic) in LMSE are not comparable to others. 

\subsection{Results on IIW} 
\label{subsec:iiw}

The IIW dataset differs substantially from MPI-Sintel and
MIT Intrinsic in its ground truth. IIW does not have dense pixel-wise ground truth with physical grounds, but only sparse pair-wise annotations of points
in each image. Therefore the losses relying on dense prediction, e.g., GAN discriminator loss/VGG loss,
in ~\cite{cheng2018intrinsic,lettry2018darn,ma2018single,shi2017learning} cannot apply. That also prevents using our spectral
loss and sub-branch losses. For this experiment, our proposed
loss functions were replaced with the WHDR loss used in~\cite{fan2018revisiting}. Our network structure remains encoded with spectral weights. 

The WHDR scores are shown in Tab.~\ref{table:iiw}. FFI-Net, despite being disabled with spectral banding and loss, still achieves competitive WHDR score $15.81$ among the state-of-the-arts~\cite{cgintrinsic,fan2018revisiting,zhou2019glosh}. The best performance by~\cite{fan2018revisiting} can be partly explained by its tailored domain filter that flattens albedo prediction. While ~\cite{cgintrinsic} and \cite{zhou2019glosh} use additional datasets (\eg CGI, SAW, SUNCG) along with IIW for training augmentation. 

\input{tab-iiw.tex}

\input{fig-iiw.tex}

\input{fig-wospecloss.tex}

\subsection{Ablation Study}
\label{secsec:results_spectralweights}
\para{Spectral Bands.}
Tab.~\ref{tab:sintel_scene_split} demonstrates that adding more spectral bands ($M$) consistently improves the results. We use $M=5$ as a trade-off between efficiency and accuracy.
It is worth mentioning that using a single spectral band already outperforms the state-of-the-arts in si-MSE and si-LMSE. We believe this is due to the spectral encoding of residual blocks and the spectral loss.

\medskip 

\para{Spatial vs Spectral Residual Blocks.}
Tab.~\ref{tab:sintel_scene_split} shows using spectral residual blocks performs significantly better than the spatial residual blocks ($R_\text{spat}$ v.s. $R_\text{spec}$). 
Fig.~\ref{figure:without_spectralloss} shows using spectral residual blocks significantly accelerates the convergence rate (3x - 5x) and improves the accuracy. 

\medskip 
\para{Spectral Loss.} Tab.~\ref{tab:sintel_scene_split} reports the results of FFI-Net without spectral loss (w/o $\mathcal L_\text{spec}$): the performance is inferior to FFI-Net on the metrics si-LMSE and DSSIM but remains the same on si-MSE. Unsurprisingly, si-MSE defines the pixel-wise loss while si-LMSE and DSSIM consider the window-wise correlations.
The results demonstrate the effectiveness of spectral loss on the global similarity between the intrinsic prediction and the corresponding ground truth. Moreover, we point out that FFI-Net (only $\mathcal L_\text{spec}$) already outperforms the state-of-the-arts~\cite{cheng2018intrinsic,lettry2018darn,fan2018revisiting} on si-MSE and si-LMSE. 

\begin{table}[t]
\setlength{\tabcolsep}{1.5pt}
\centering
\small
\begin{tabular}{l ccccc}
\hline
{Method}& Fan \cite{fan2018revisiting} & Darn \cite{lettry2018darn} & Lap \cite{cheng2018intrinsic} & CGI~\cite{cgintrinsic} & FFI-Net \cr
\hline
Parameters(Train) ~ & 2.80 &  17.97 & 138 & 7.7 & 0.75 \cr
Parameters(Test) ~ & 2.80 &  0.74 & 0.94 & 7.7 & 0.75  \cr
\hline
\vspace{-5mm}
\end{tabular}
\caption{Network parameter numbers displayed in million-scale. The training parameters for \cite{lettry2018darn} and \cite{cheng2018intrinsic} includes that of the GAN discriminator and the VGG classifier, respectively.
}
\label{tab:parameter_and_time}. 
\end{table}

\medskip 
\para{Network parameters and inference time. }
Tab.~\ref{tab:parameter_and_time} profiles the network parameter number of FFI-Net. 
It generally needs less parameters compared to the others, which use extra guidance network~\cite{fan2018revisiting}, adversarial network~\cite{lettry2018darn}, VGGNet~\cite{cheng2018intrinsic} and U-Net~\cite{cgintrinsic} for training. 
\miaojing{As for inference time, we reproduced~\cite{fan2018revisiting,cheng2018intrinsic} with the same GPU (GTX 1080) as ours, it is $\sim$100 ms for~\cite{fan2018revisiting}, $\sim$70 ms for~\cite{cheng2018intrinsic} and $\sim$42 ms for FFI-Net.}



%% file: tab_mpi_scenesplit.tex
\setlength{\tabcolsep}{1.5pt} 
{
\begin{table}[t]
\centering
\small
\resizebox{0.99\linewidth}{!}{
\begin{tabular}{l rrr | rrr | rrr}
\toprule
{Sintel \textit{scene split}}&
\multicolumn{3}{c}{si-MSE $\times10^{2}$}&\multicolumn{3}{c}{si-LMSE$\times10^{2}$}&\multicolumn{3}{c}{DSSIM$\times10^{2}$}\cr\cline{2-10}
&A&S&avg&A&S&avg&A&S&avg\cr\hline
DI~\cite{narihira2015direct}&2.01&2.24&2.13&1.31&1.48&1.39&20.73&15.94&18.33\cr
DARN~\cite{lettry2018darn}&1.77&1.84&1.81&0.98&0.95&0.97&14.21&14.05&14.13\cr
Fan et al.~\cite{fan2018revisiting}&1.81&1.75&1.78&1.22&1.18&1.20&16.74&13.82&15.28\cr
LapPyrNet~\cite{cheng2018intrinsic}& 1.38 & 1.38 & 1.38 & 0.92 & 0.93 & 0.92 & \textbf{8.46} & 9.26 & \cellcolor{gray!25} 8.86 \cr
FFI-Net & \textbf{1.19} &  \textbf{1.16} & \textbf{1.18} & \textbf{0.73} & \textbf{0.73} & \textbf{0.73} &  \cellcolor{gray!25}8.93 & \textbf{8.60} & \textbf{8.77} \cr
\hline
M=5,$R_{\text{spec}}$ & \textbf{1.19} &  \textbf{1.16} & \textbf{1.18} & \textbf{0.73} & \textbf{0.73} & \textbf{0.73} &  \cellcolor{gray!25}8.93 & \textbf{8.60} & \textbf{8.77} \cr
M=3,$R_{\text{spec}}$ & 1.24 &  1.16 & 1.20 & 0.75 & 0.75 & 0.75 &  9.59 & 8.85  & 9.22\\
M=1,$R_{\text{spec}}$ & 1.35 &  1.36 & 1.35 & 0.82 & 0.82 & 0.82 &  10.43 & 10.84 & 10.63\\
M=5,$R_{\text{spat}}$ & 1.50 &  1.50 & 1.50~& 0.91 & 0.92 & 0.91 &  10.25 & 9.78  & 10.01\\
M=3,$R_{\text{spat}}$ & 1.41 &  1.44 & 1.42~& 0.82 & 0.93 & 0.88 &  10.59 & 9.95  & 10.27\\
M=1,$R_{\text{spat}}$ & 1.52 & 1.67 & 1.60&~0.91 & 1.14 & 1.02 & 11.04 & 11.69 & 11.37\\
\hline
w/o $\mathcal L_{\text{spec}}$ (M =5, $R_{\text{spec}}$) & \textbf{1.19} &  \textbf{1.16} & \textbf{1.18} & \cellcolor{gray!25}0.75 & 0.75 & \cellcolor{gray!25}0.75 &  9.55 &\cellcolor{gray!25}8.70 & 9.13 \cr
only $\mathcal L_{\text{spec}}$ (M =1, $R_{\text{spat}}$) & 1.29 &  1.28 & 1.29 & 0.77 & 0.84 & 0.81 &  10.11 & 9.13 & 9.62 \cr

\bottomrule
\vspace{-3mm}
\end{tabular}
}
\caption{Results for MPI-Sintel (scene split). si-MSE, si-LMSE and DSSIM are reported for albedo (A), shading (S) and their average (avg). We also ablate the key elements of FFI-Net: number of spectral bands $M~\in~\{1,3,5\}$,  spectral ($R_\text{spec}$) vs. spatial ($R_\text{spat}$) residual blocks, 
and the spectral loss~$\mathcal L_{\text{spec}}$. FFI-Net is by default with M=5, $R_{\text{spec}}$.
The best results are marked in bold.
} 
\label{tab:sintel_scene_split}. 
\vspace{-5mm}
\end{table}
}

%% file: fig-mpi-scenesplit.tex
\setlength{\tabcolsep}{1pt}
\renewcommand{\arraystretch}{1}
\begin{figure}[t]
\begin{center}

\begin{tabular}{ccc}

\vspace{-0.75mm}
\raisebox{.9\height}{\rotatebox{90}{\scriptsize{{Input}}}}
&\multicolumn{2}{c}{\includegraphics[width=3.6cm]{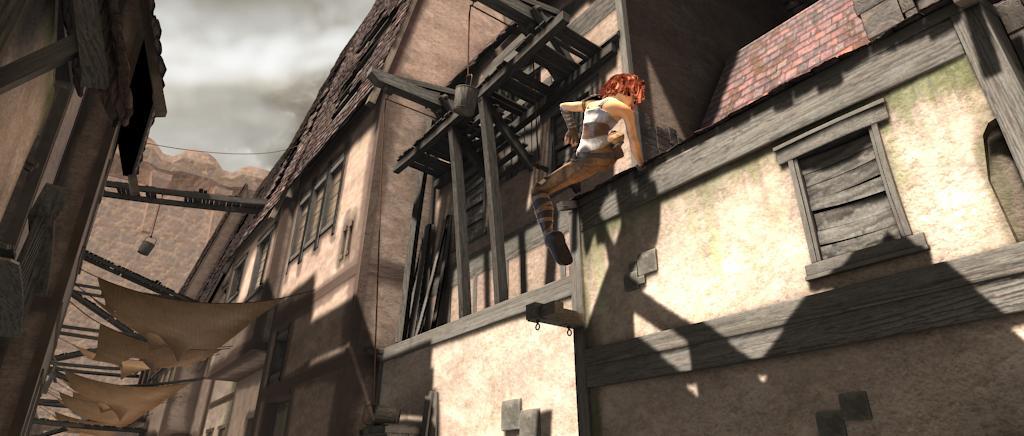}}
\\
& (A) & (S) \\
\vspace{-0.75mm}
\raisebox{1\height}{\rotatebox{90}{\scriptsize{{DI~\cite{narihira2015direct}}}}}
&{{\includegraphics[width=3.6cm]{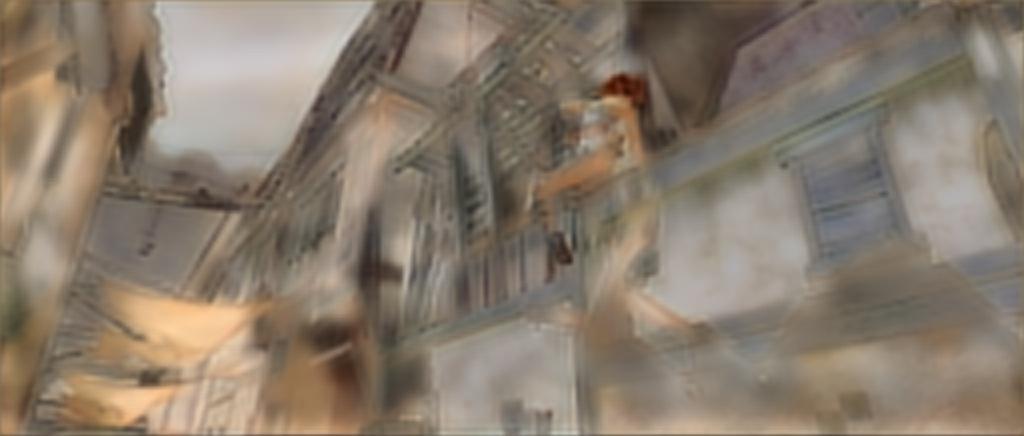}}}
&{{\includegraphics[width=3.6cm]{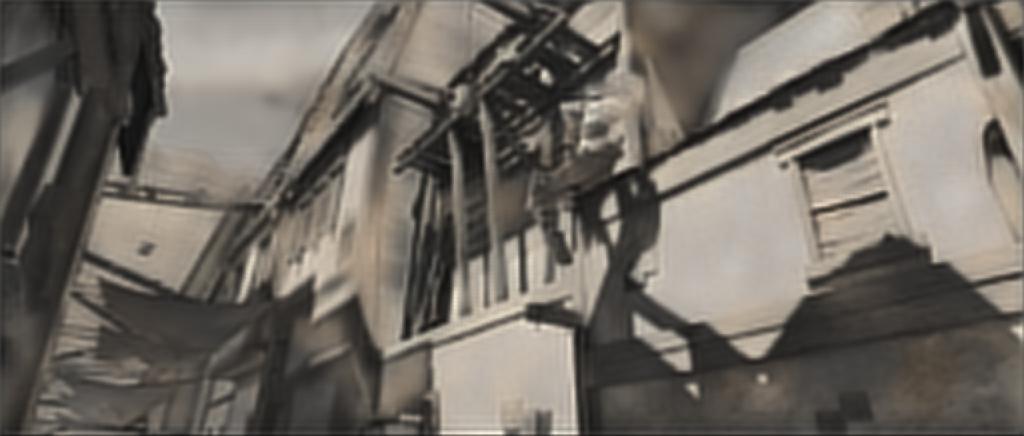}}}
\\

\vspace{-0.75mm}
\raisebox{1\height}{\rotatebox{90}{\scriptsize{{Fan \cite{fan2018revisiting}}}}}
&{{\includegraphics[width=3.6cm]{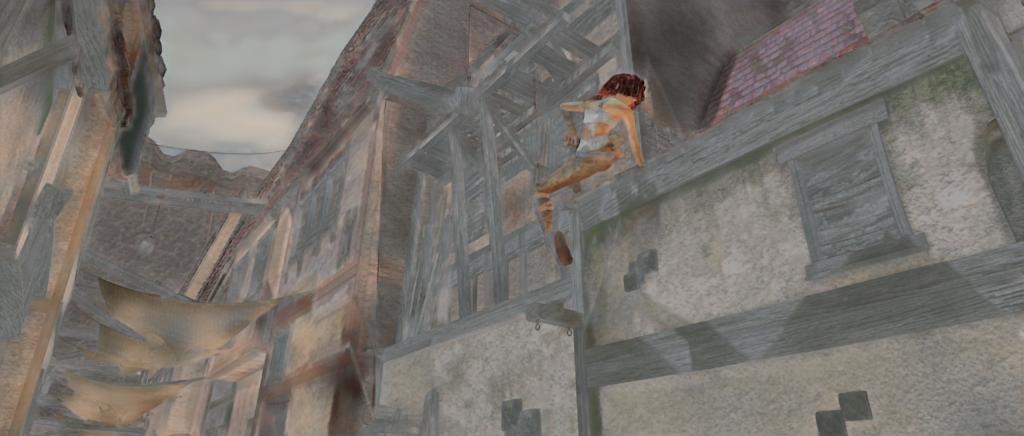}}}
&{{\includegraphics[width=3.6cm]{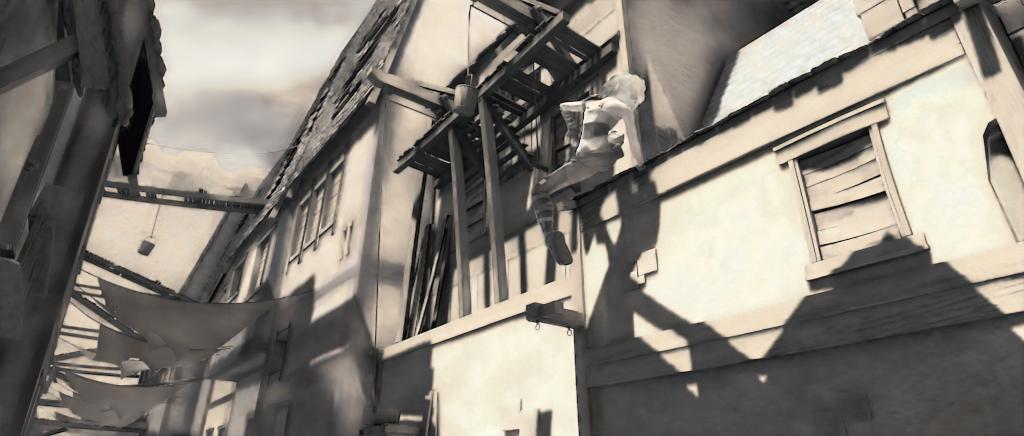}}}
\\

\vspace{-0.75mm}
\raisebox{1\height}{\rotatebox{90}{\scriptsize{{~~~~FFI}}}}
&{{\includegraphics[width=3.6cm]{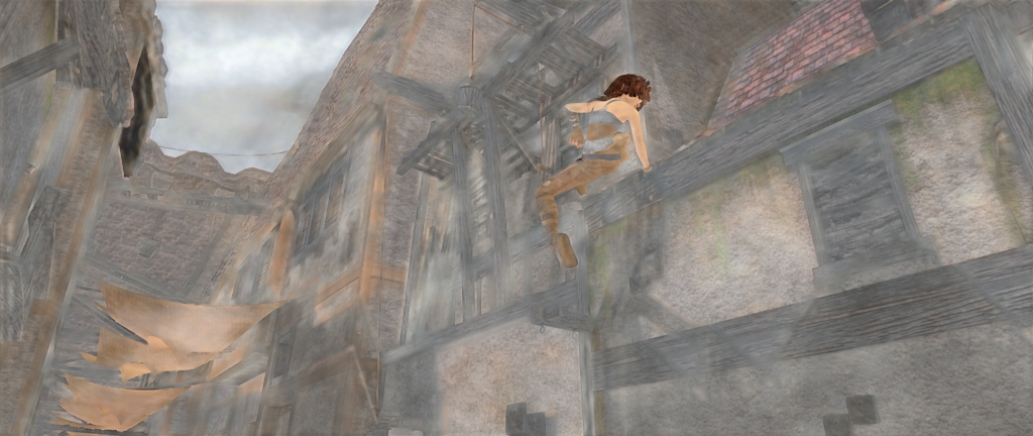}}}
&{{\includegraphics[width=3.6cm]{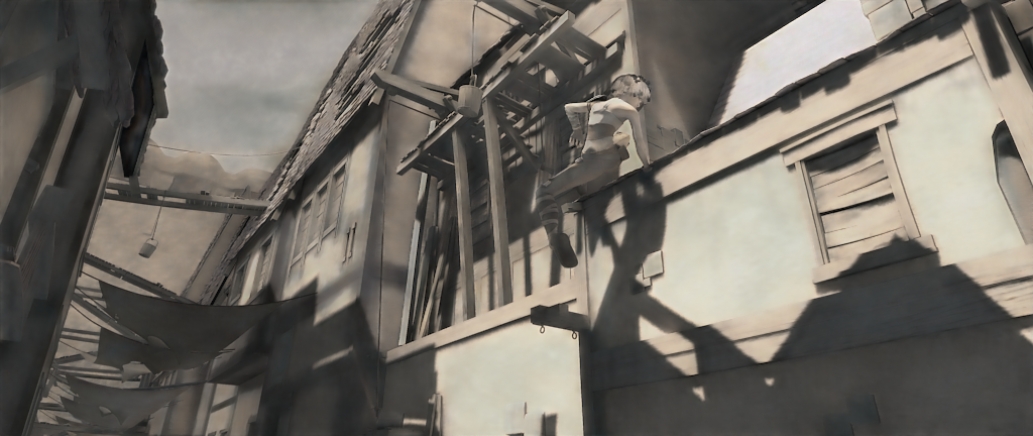}}}
\\

\vspace{-0.75mm}
{\rotatebox{90}{\scriptsize{{Ground Truth}}}}
&{{\includegraphics[width=3.6cm]{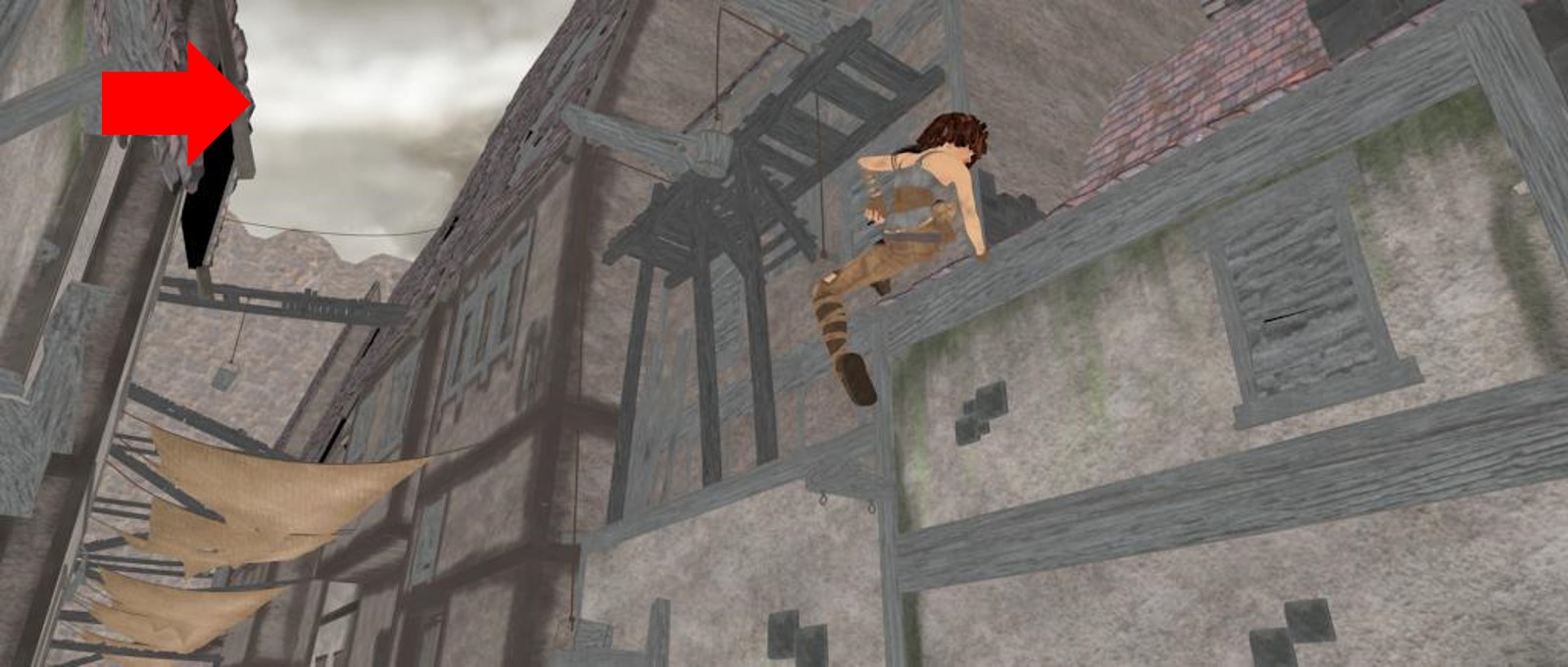}}}
&{{\includegraphics[width=3.6cm]{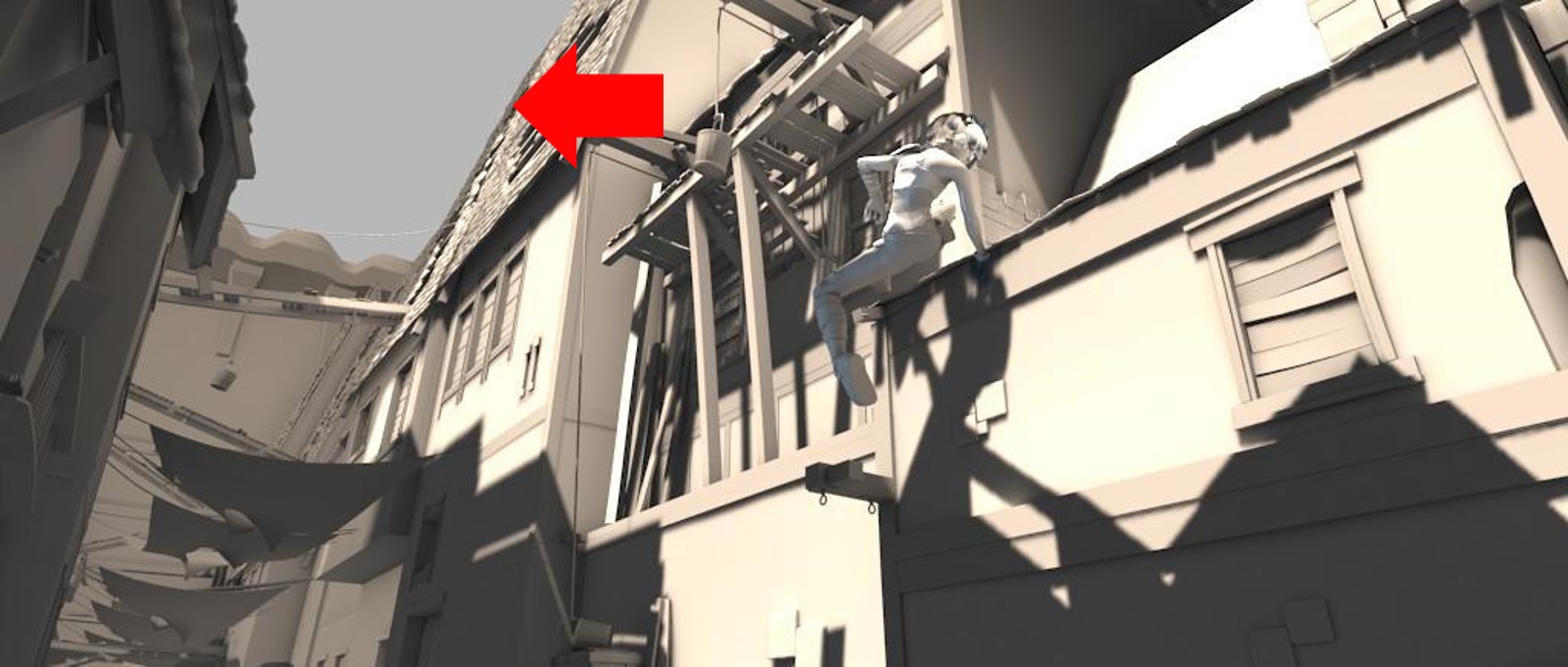}}}
\\

\end{tabular}

\end{center}
\caption{Examples on the MPI-Sintel (scene split). We emphasize the ability of FFI-Net of shadow removal (\eg sky - red arrows) in the unseen scene compared to others. 
}
\label{figure:mpi_scenesplit}
\end{figure}


%% file: tab_mpi_imagesplit.tex
\setlength{\tabcolsep}{1.5pt}
{
\begin{table}[t]
\centering
\resizebox{0.99\linewidth}{!}{
\small
\begin{tabular}{l rrr | rrr | rrr}
\toprule
{Sintel \textit{image split}}&
\multicolumn{3}{c}{si-MSE$\times10^{2}$}&\multicolumn{3}{c}{si-LMSE$\times10^{2}$}&\multicolumn{3}{c}{DSSIM$\times10^{2}$}\\
&A&S&avg&A&S&avg&A&S&avg\\
\midrule
Retinex \cite{grosse09intrinsic} ~&6.06&7.27&6.67&3.66&4.19&3.93&22.70&24.00&23.35\cr
Lee et al. \cite{lee2012estimation}~&4.63&5.07&4.85&2.24&1.92&2.08&19.90&17.70&18.80\cr
SIRFS~ \cite{barron2015shape}&4.20&4.36&4.28&2.98&2.64&2.81&21.00&20.60&20.80\cr
Chen et al.~\cite{chen2013simple}&3.07&2.77&2.92&1.85&1.90&1.88&19.60&16.50&18.05\cr
DI~\cite{narihira2015direct}&1.00&0.92&0.96&0.83&0.85&0.84&20.14&15.05&17.60\cr
DARN~\cite{lettry2018darn}&1.24&1.28&1.26&0.69&0.70&0.70&12.63&12.13&12.38\cr
Kim et al.~\cite{kim2016unified}&0.70&0.90&0.70&0.60&0.70&0.70&9.20&10.10&9.70\cr
Fan et al.~\cite{fan2018revisiting} &0.67& \cellcolor{gray!25}0.60& \cellcolor{gray!25}0.63& \textbf{0.41}& \cellcolor{gray!25}0.42& \textbf{0.41} &10.50&7.83&9.16\cr
LapPyrNet~\cite{cheng2018intrinsic}& \cellcolor{gray!25}0.66 & \cellcolor{gray!25}0.60 & \cellcolor{gray!25}0.63 & \cellcolor{gray!25}0.44 & \cellcolor{gray!25}0.42 & 0.43 & {\bf 6.56} & \cellcolor{gray!25}{6.37} & \textbf{6.47} \cr
Liu \etal \cite{liu2020cvpr} & 1.59 & 1.48 & 1.54 & 0.87 & 0.81 & 0.84 & 17.97 & 14.74 & 16.35 \cr
\midrule
FFI-Net ~& \textbf{0.64} & \textbf{0.57} & \textbf{0.61} & 0.46 & \textbf{0.39} &  \cellcolor{gray!25}{0.42} & \cellcolor{gray!25}7.73 &  \textbf{6.28} & \cellcolor{gray!25}7.01 \cr
\bottomrule
\end{tabular}}

\caption{Results for MPI-Sintel (image split). 
}
\vspace{-5mm}
\label{tab:sintel_image_split}.
\end{table}
}

%% file: fig-mpi-imagesplit.tex
\begin{figure}[t]
\begin{center}

\begin{tabular}{ccc}

\vspace{-0.75mm}
\raisebox{.9\height}{\rotatebox{90}{\scriptsize{{Input}}}}
&\multicolumn{2}{c}{\includegraphics[width=3.6cm]{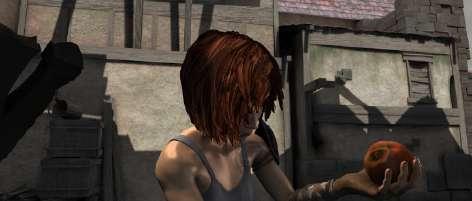}}
\\
& (A) & (S) \\
\vspace{-0.75mm}
\raisebox{0.5\height}{\rotatebox{90}{\scriptsize{{Lee~\cite{lee2012estimation}}}}}
&{{\includegraphics[width=3.6cm]{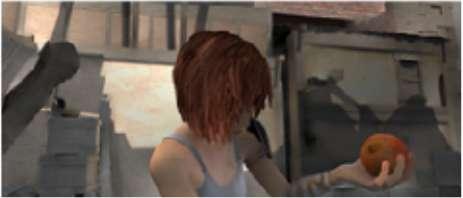}}}
&{{\includegraphics[width=3.6cm]{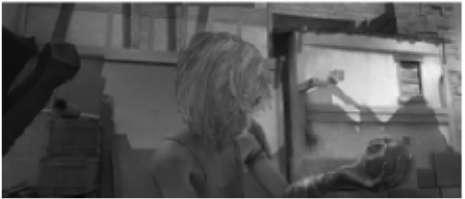}}}
\\

\vspace{-0.75mm}
\raisebox{0.5\height}{\rotatebox{90}{\scriptsize{{Barron~\cite{barron2015shape}}}}}
&{{\includegraphics[width=3.6cm]{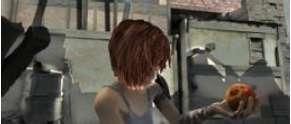}}}
&{{\includegraphics[width=3.6cm]{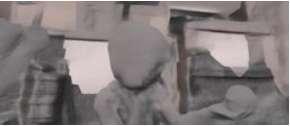}}}
\\

\vspace{-0.75mm}
\raisebox{0.5\height}{\rotatebox{90}{\scriptsize{{Chen~\cite{chen2013simple}}}}}
&{{\includegraphics[width=3.6cm]{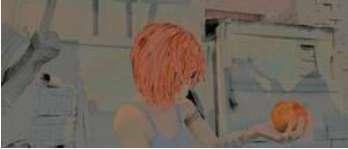}}}
&{{\includegraphics[width=3.6cm]{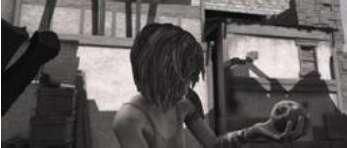}}}
\\

\vspace{-0.75mm}
\raisebox{1\height}{\rotatebox{90}{\scriptsize{{DI~\cite{narihira2015direct}}}}}
&{{\includegraphics[width=3.6cm]{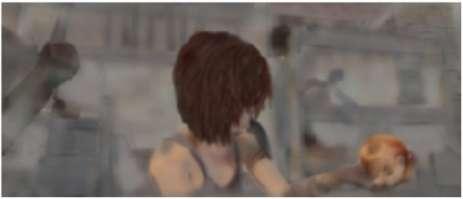}}}
&{{\includegraphics[width=3.6cm]{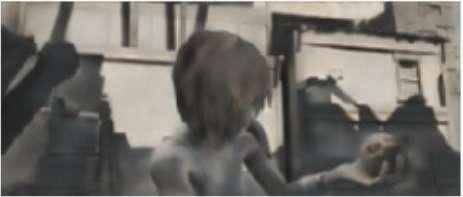}}}
\\

\vspace{-0.75mm}
\raisebox{1\height}{\rotatebox{90}{\scriptsize{{Fan \cite{fan2018revisiting}}}}}
&{{\includegraphics[width=3.6cm]{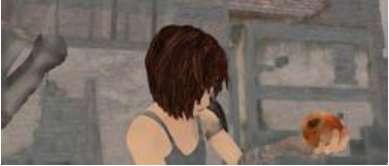}}}
&{{\includegraphics[width=3.6cm]{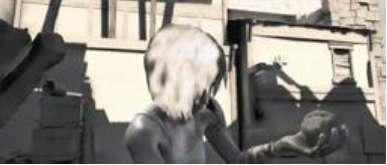}}}
\\

\vspace{-0.75mm}
\raisebox{1\height}{\rotatebox{90}{\scriptsize{{Lap\cite{cheng2018intrinsic}}}}}
&{{\includegraphics[width=3.6cm]{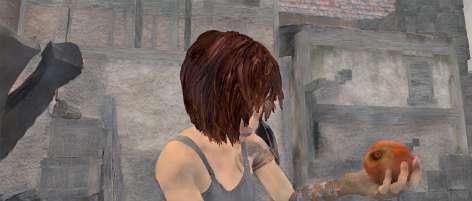}}}
&{{\includegraphics[width=3.6cm]{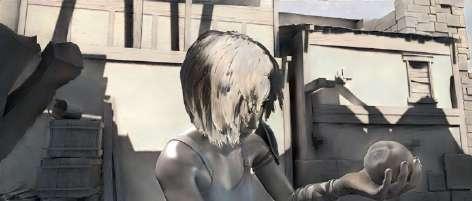}}}
\\

\vspace{-0.75mm}
\raisebox{1\height}{\rotatebox{90}{\scriptsize{{~~~~FFI}}}}
&{{\includegraphics[width=3.6cm]{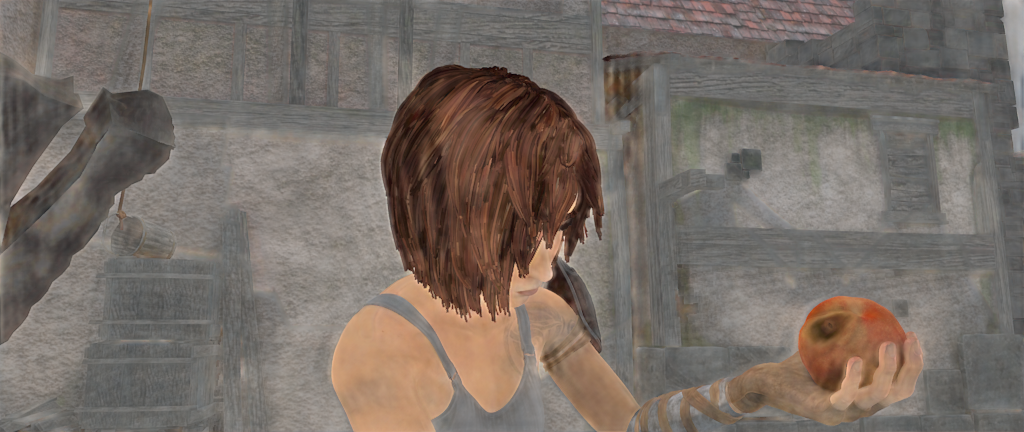}}}
&{{\includegraphics[width=3.6cm]{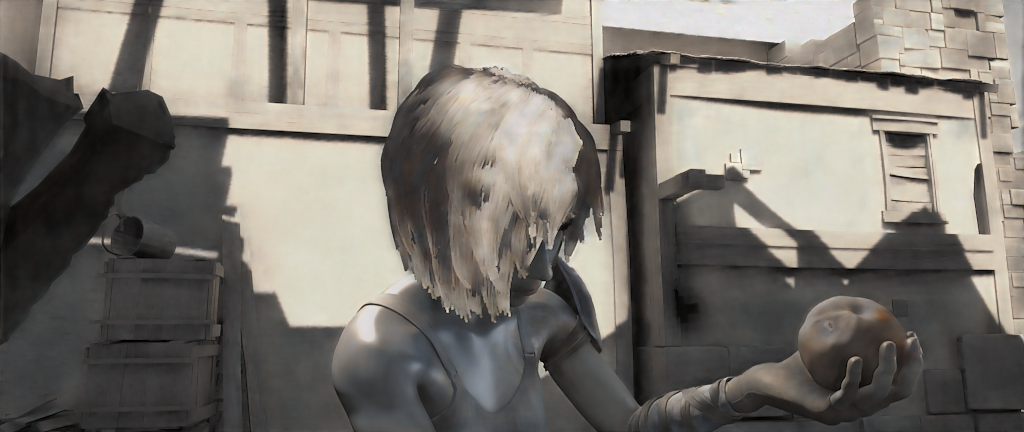}}}
\\

\vspace{-0.75mm}
{\rotatebox{90}{\scriptsize{{Ground Truth}}}}
&{{\includegraphics[width=3.6cm]{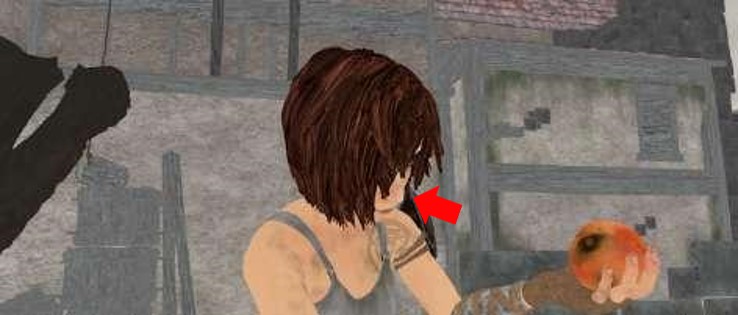}}}
&{{\includegraphics[width=3.6cm]{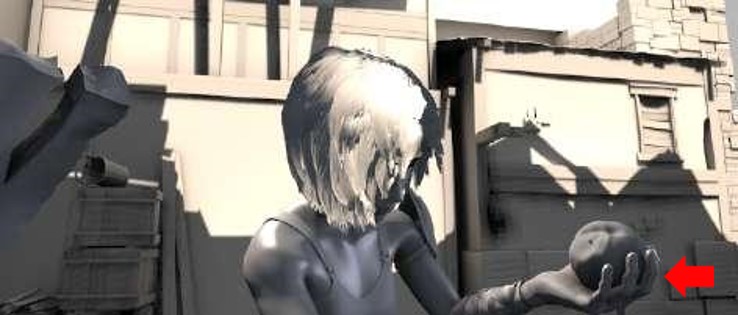}}}
\\
\vspace{-5mm}

\end{tabular}

\end{center}
\vspace{-5mm}
\caption{Examples of (A)lbedo and (S)hading predictions on MPI-Sintel (image split). FFI-Net performs well in difficult parts, \eg person fingers, faces (red arrows).
}
\label{figure:mpi_imagesplit}
\vspace{-2mm}
\end{figure}


%% file: tab_mit_A.tex
\begin{table}
\setlength{\tabcolsep}{2pt} 
\renewcommand{\arraystretch}{1} 
\centering
\small
\begin{tabular}{l c cccc}
\toprule
~ & Tr. &
\multicolumn{3}{c}{si-MSE $\times10^{2}$}&\multicolumn{1}{c}{si-LMSE$\times10^{2}$} \cr\cline{3-6}
&data&A&S&avg&Total\cr
\midrule
SIRFS \cite{barron2015shape} & M & 0.64 & 0.98 & 0.81 & 1.25 \cr
\midrule
Zhou \textit{et al.} \cite{zhou2015learning} & M & 2.52 & 2.29 & 2.40 & 3.19 \cr
Shi \textit{et al.} \cite{shi2017learning} & SN+M & 2.16 & 1.35 & 1.75 & 2.71 \cr
DI~\cite{narihira2015direct} & ST+M & 2.07 & 1.24 & 1.65 & 2.39 \cr


Fan et al.~\cite{fan2018revisiting}& M &  1.34 & 0.89 & 1.11 & 2.03 \cr
LapPyrNet~\cite{cheng2018intrinsic}& I+M &  0.89 & \textbf{0.73} & 0.81 & 1.41  \cr
FFI-Net  ~ & M & \textbf{0.71} &  0.76 & \textbf{0.73} &  \textbf{1.31} \cr 
\bottomrule
\end{tabular}
\caption{
Results for MIT Intrinsic. All methods use the training-test split file released in~\cite{barron2015shape}. 
For si-MSE and si-LMSE, we use the evaluation code from \cite{fan2018revisiting,barron2015shape}. Tr. data lists the datasets used in training: MIT (M), ShapeNet (SN), Sintel (ST) and ImageNet (I). The best CNN-based results are marked in bold. 
}. 
\label{tab:mit}
\vspace{-2ex}
\end{table}

%% file: tab_mit_B.tex
\begin{table}
\setlength{\tabcolsep}{3pt} 
\renewcommand{\arraystretch}{1} 
\centering
\small
\begin{tabular}{l c cccccc}
\hline
~ & Tr. &
\multicolumn{2}{c}{MSE} &\multicolumn{2}{c}{LMSE} &\multicolumn{2}{c}{DSSIM}\cr
\cline{3-8} & data &A&S &A&S &A&S \cr\hline
SIRFS~\cite{barron2015shape} & M & 1.47 & 0.83 & 4.16 & 1.68 & 12.38 &{9.85} \cr
\midrule 
DI~\cite{narihira2015direct} & ST+M & 2.77 & 1.54 & 5.86 & 2.95 & 15.26 &13.28 \cr
Shi \textit{et al.} \cite{shi2017learning} & SN+M & 2.78 & 1.26 & 5.03 & 2.40 & 14.65 & 12.00\cr
CGI \cite{cgintrinsic} & CGI & 2.21 & 1.86 & 3.49 & 2.59 & 17.39 & 16.52 \cr
CGI \cite{cgintrinsic} & CGI+M & {1.67} & 1.27 & 3.19 & 2.21 & 12.87 & 13.76 \cr
Ma \textit{et al.} \cite{ma2018single}$^*$ & M  & 1.68 & {0.93} & {\em 0.74} & {\em 0.52} & -- & -- \cr
Liu \textit{et al.} \cite{liu2020cvpr}
& M & 1.57 & {1.35} & {1.46} & {2.31} & -- & -- \cr
FFI-Net & M & \textbf{1.11} & \textbf{0.93} & { 2.91} & 3.19 & \textbf{10.14} & \textbf{11.39} \cr
\hline
\end{tabular}
\caption{
More results for MIT Intrinsic under diffferent metrics, MSE, LMSE, and DSSIM (\cite{ma2018single}$^*$ uses a different window size for LMSE). The best CNN-based results are marked in bold. 
}
\label{tab:mit_b}
\vspace{-1ex}
\end{table}

%% file: fig-mit.tex
\begin{figure}[htp]
\begin{center}
\resizebox{1.0\linewidth}{!}{
\begin{tabular}{c c ccccc}
\vspace{-0.5mm}
\!\!\includegraphics[width=1.4cm]{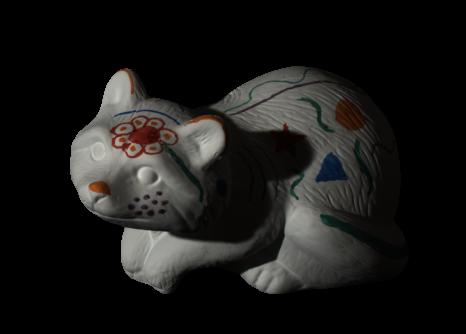}
& \raisebox{3mm}{\rotatebox{0}{(A)}}
&\includegraphics[width=1.4cm]{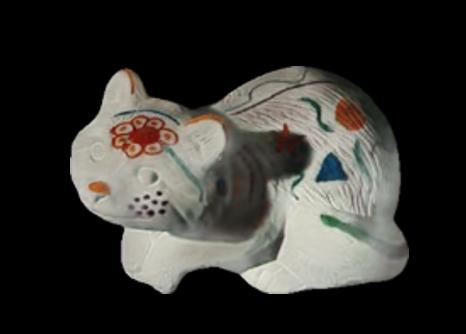}
&\includegraphics[width=1.4cm]{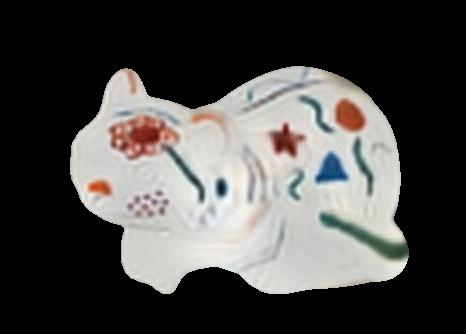}
&\includegraphics[width=1.4cm,height=1.003cm]{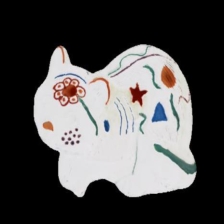}
&\includegraphics[width=1.4cm,,height=1.003cm]{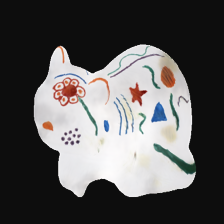}
&\includegraphics[width=1.4cm]{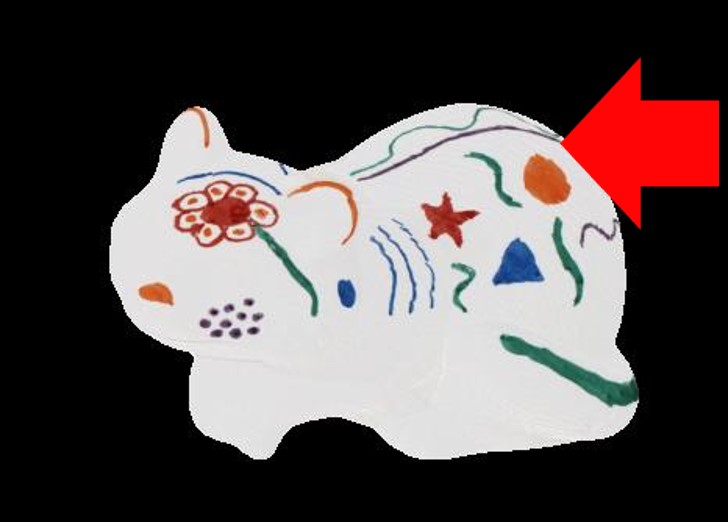}\cr
& \raisebox{3mm}{\rotatebox{0}{(S)}}
&\includegraphics[width=1.4cm]{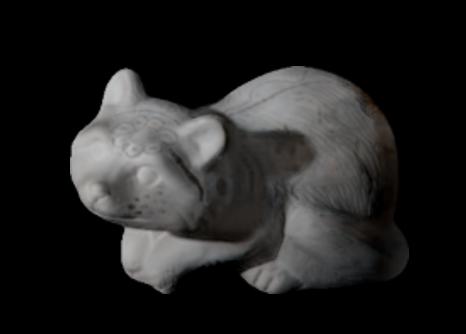}
&\includegraphics[width=1.4cm]{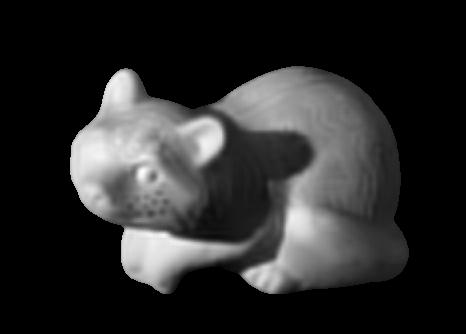}
&\includegraphics[width=1.4cm,,height=1.003cm]{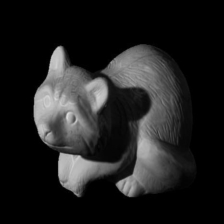}
&\includegraphics[width=1.4cm,height=1.003cm]{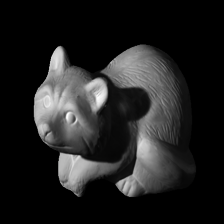}
&\includegraphics[width=1.4cm]{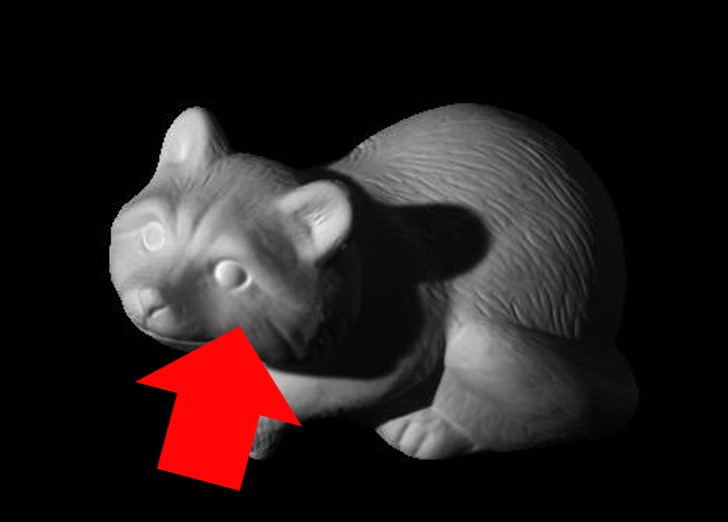}\\

\vspace{-0.5mm}

\!\!\includegraphics[width=1.4cm]{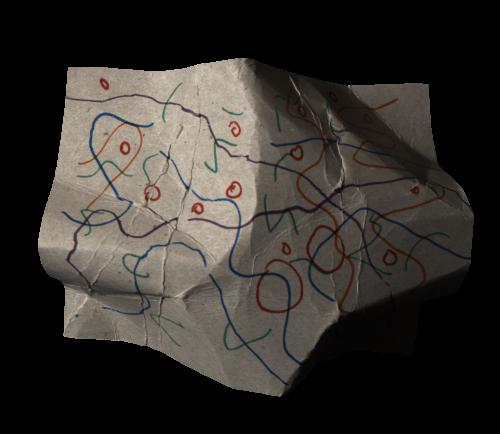}
& \raisebox{3mm}{\rotatebox{0}{(A)}}
&\includegraphics[width=1.4cm]{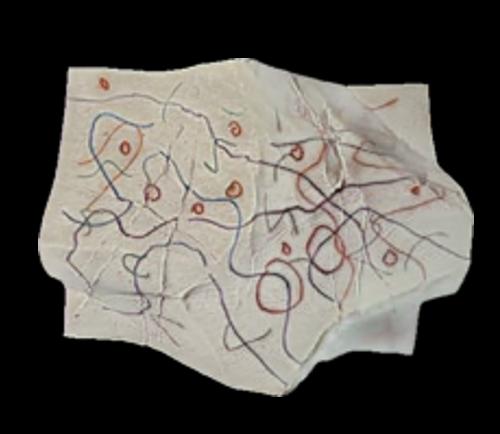}
&\includegraphics[width=1.4cm]{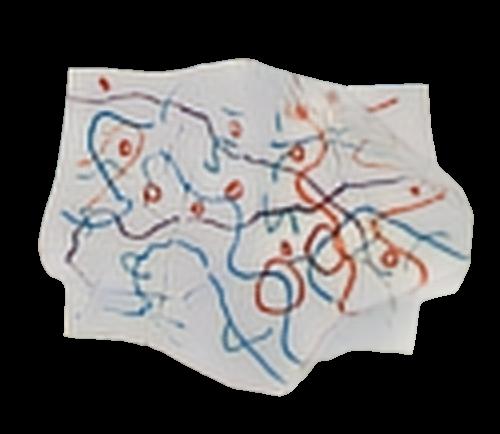}
&\includegraphics[width=1.4cm, height=1.215cm]{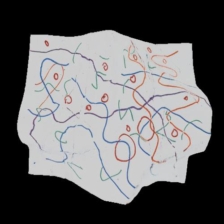}
&\includegraphics[width=1.4cm,,height=1.215cm]{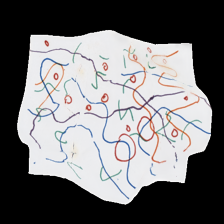}
&\includegraphics[width=1.4cm]{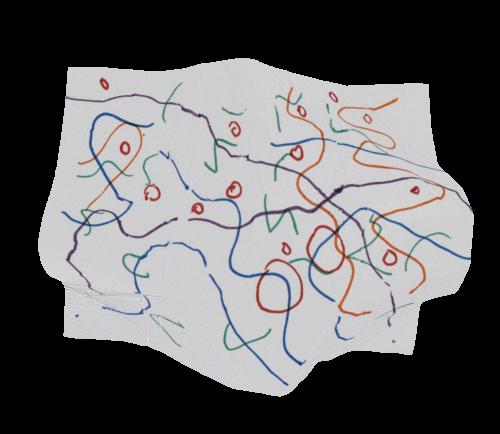}\\
& \raisebox{3mm}{\rotatebox{0}{(S)}}
&\includegraphics[width=1.4cm]{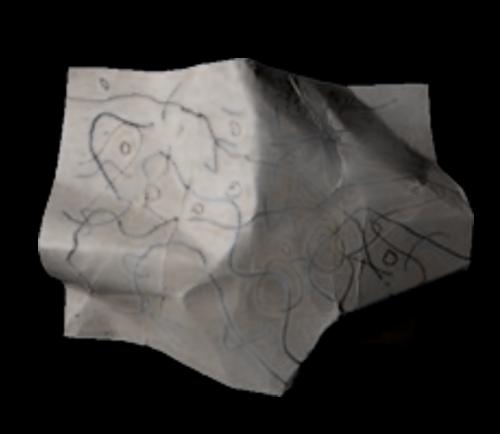}
&\includegraphics[width=1.4cm]{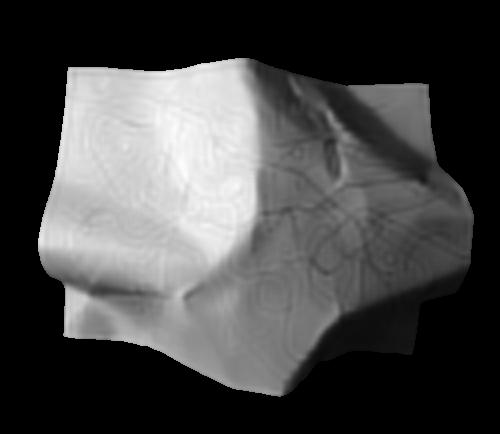}
&\includegraphics[width=1.4cm,,height=1.215cm]{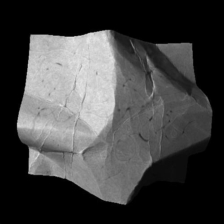}
&\includegraphics[width=1.4cm, height=1.215cm]{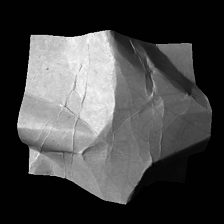}
&\includegraphics[width=1.4cm]{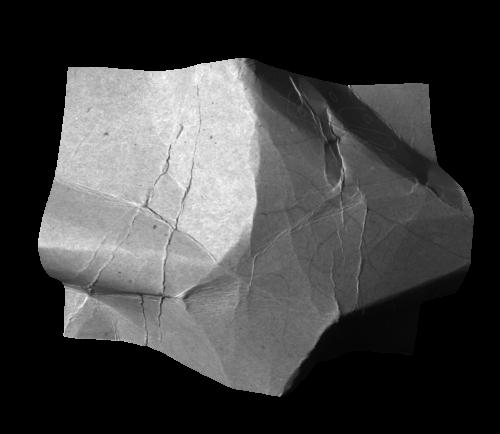}\\

%
%
%

%

\vspace{-0.5mm}

\small{Input} & & \small{Shi \emph{et al.} \cite{shi2017learning}} & \small{DI \cite{narihira2015direct}} & \small{Fan \textit{et al.} \cite{fan2018revisiting}} & \small{FFI-Net} & \small{Ground Truth}\\


\end{tabular}}

\end{center}
\vspace{-2mm}
\caption{Sample (A)lbedo and (S)hading on MIT Intrinsic. FFI-Net performs better for object details (top), as well as for the overall appearance (bottom). 
}
\label{figure:mit}
\end{figure}

%% file: tab-iiw.tex
\begin{table}
\begin{center}
	\small
\begin{tabular}{llc}
\toprule
Methods & Training set &  WHDR (mean) \\
\midrule
DI \cite{narihira2015direct} & ST & 37.30 \\
Shen \emph{et al.} \cite{shen2011intrinsic} & -- & 36.90 \\
Retinex (color) \cite{grosse2009ground} & -- & 26.89 \\
Retinex (gray) \cite{grosse2009ground} & -- & 26.84 \\
Garces \emph{et al.} \cite{garces2012intrinsic} & --& 25.46 \\
Zhao \emph{et al.} \cite{zhao2012closed} & -- & 23.20 \\
$L_1$ flattening \cite{bi20151} & -- & 20.94 \\
Bell \emph{et al.} \cite{bell2014intrinsic} & -- & 20.64 \\
Zhou \emph{et al.} \cite{zhou2015learning} & IIW & 19.95 \\
Nestmeyer \emph{et al.} CNN\cite{nestmeyer2017reflectance} & IIW & 19.49 \\
Zoran \emph{et al.} * \cite{zoran2015learning} & IIW & 17.85\\
Nestmeyer \emph{et al.} \cite{nestmeyer2017reflectance} & IIW & 17.69 \\
Bi \emph{et al.} \cite{bi20151}& -- & 17.67 \\
CGI \cite{cgintrinsic} & CGI+IIW & 17.50 \\
CGI \cite{cgintrinsic} & CGI+IIW(A)+SAW & 15.50 \\
Fan \emph{et al.} \cite{fan2018revisiting} & IIW & \textbf{14.45} \\
Zhou \etal \cite{zhou2019glosh} & SUNCG + IIW & 15.20 \\
Liu \etal~\cite{liu2020cvpr}  & IIW & 18.69 \\
\midrule
FFI-Net w/ WHDR loss & IIW & {15.81} \\
\bottomrule
\end{tabular}
\end{center}
\vspace{-2mm}
\caption{Results for IIW. Evaluation follows the train-test split in \cite{narihira2015direct}. Note that due to the sparse ground truth available, FFI-Net does not use the spectral banding and loss, but instead the WHDR loss in ~\cite{fan2018revisiting}.
}
\label{table:iiw}
\vspace{-0.5mm}
\end{table}

%% file: fig-iiw.tex
\begin{figure*}[ht]
  \centering
    \begin{tabular}{@{}c@{}c@{}c@{}c@{}c@{}c@{}c@{}c@{}c@{}}
        \includegraphics[width=0.11\textwidth, height=0.07\textwidth]{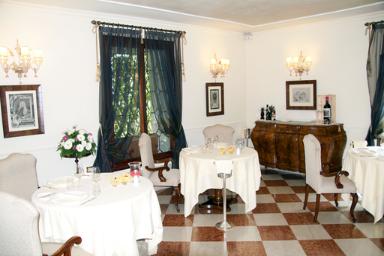}  &
        \includegraphics[width=0.11\textwidth,height=0.07\textwidth]{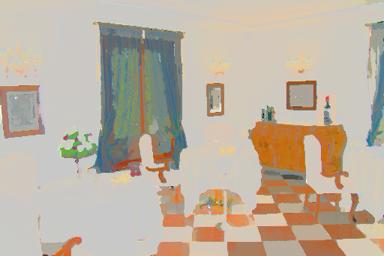}  & 
		\includegraphics[width=0.11\textwidth,height=0.07\textwidth]{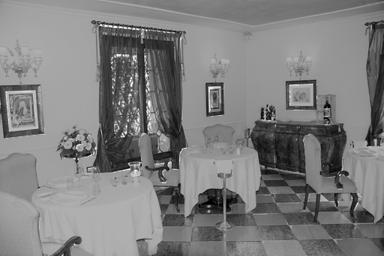}  &         
        \includegraphics[width=0.11\textwidth,height=0.07\textwidth]{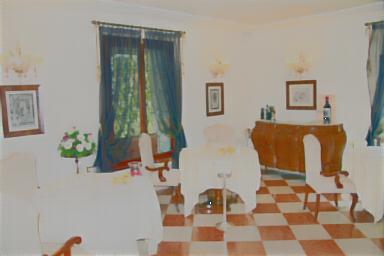}  & 
        \includegraphics[width=0.11\textwidth,height=0.07\textwidth]{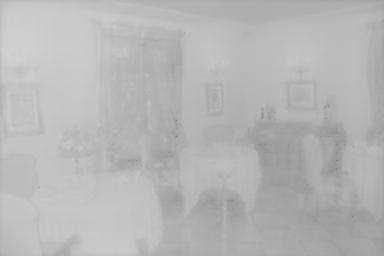}  & 
        \includegraphics[width=0.11\textwidth,height=0.07\textwidth]{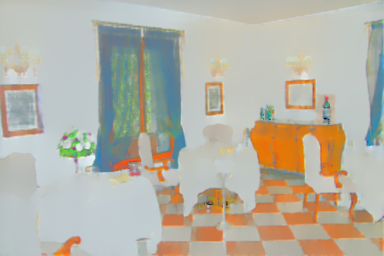}  & 
        \includegraphics[width=0.11\textwidth,height=0.07\textwidth]{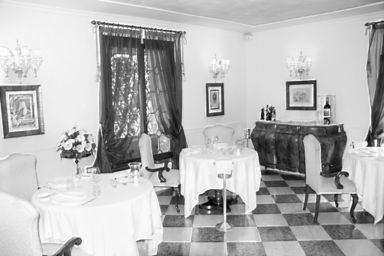}  & 
        \includegraphics[width=0.11\textwidth,height=0.07\textwidth]{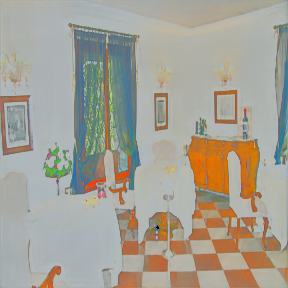}  & 
        \includegraphics[width=0.11\textwidth,height=0.07\textwidth]{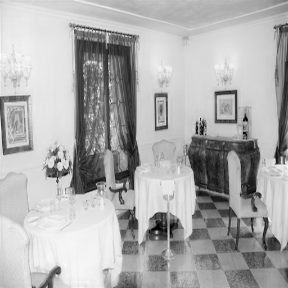}  
        
        \vspace{-1mm}
        \\  
        
        
        \includegraphics[width=0.11\textwidth,height=0.07\textwidth]{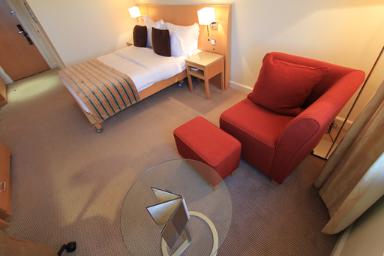} &
        \includegraphics[width=0.11\textwidth,height=0.07\textwidth]{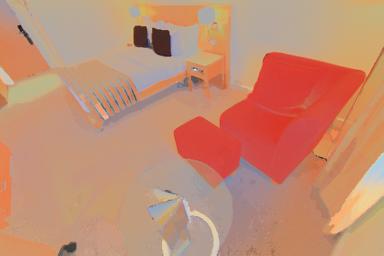} & 
		\includegraphics[width=0.11\textwidth,height=0.07\textwidth]{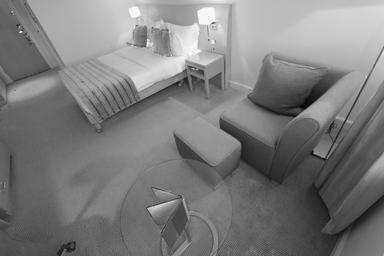} &         
        \includegraphics[width=0.11\textwidth,height=0.07\textwidth]{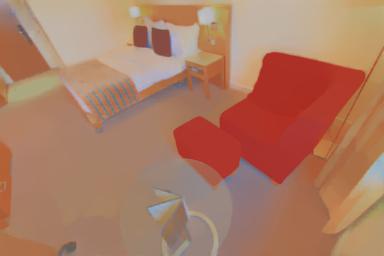}  & 
        \includegraphics[width=0.11\textwidth,height=0.07\textwidth]{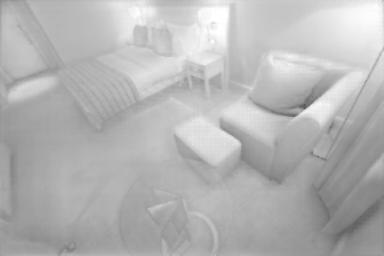} &
        \includegraphics[width=0.11\textwidth,height=0.07\textwidth]{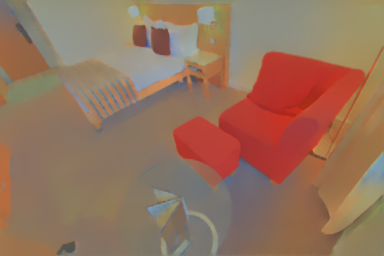}  & 
        \includegraphics[width=0.11\textwidth,height=0.07\textwidth]{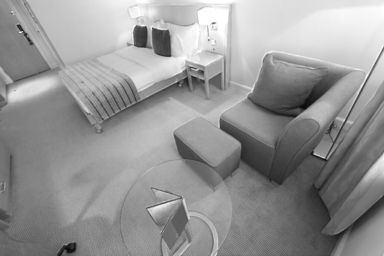} &
        \includegraphics[width=0.11\textwidth,height=0.07\textwidth]{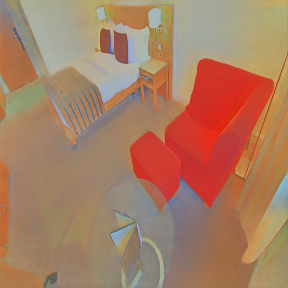}  & 
        \includegraphics[width=0.11\textwidth,height=0.07\textwidth]{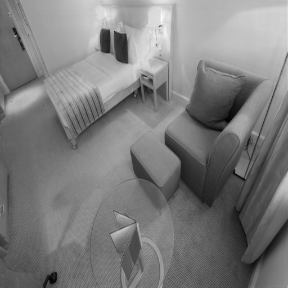}
        
        \vspace{-1mm}
        \\          

        \includegraphics[width=0.11\textwidth,height=0.07\textwidth]{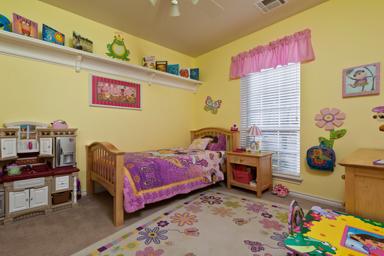}  &
        \includegraphics[width=0.11\textwidth,height=0.07\textwidth]{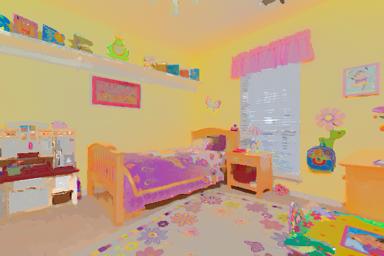} & 
		\includegraphics[width=0.11\textwidth,height=0.07\textwidth]{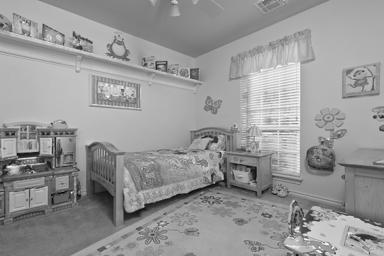}  &         
        \includegraphics[width=0.11\textwidth,height=0.07\textwidth]{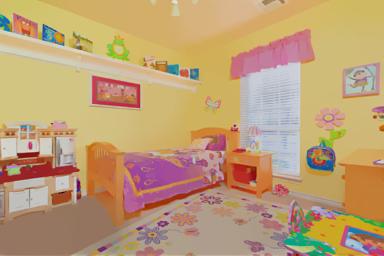}  & 
        \includegraphics[width=0.11\textwidth,height=0.07\textwidth]{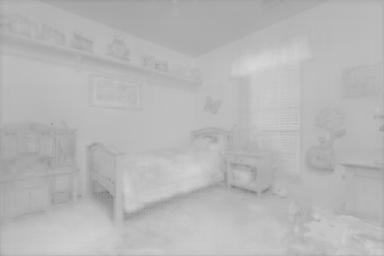}  &
        \includegraphics[width=0.11\textwidth,height=0.07\textwidth]{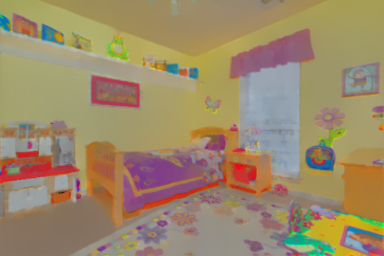}  & 
        \includegraphics[width=0.11\textwidth,height=0.07\textwidth]{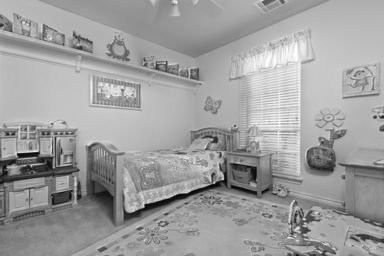}  &
        \includegraphics[width=0.11\textwidth,height=0.07\textwidth]{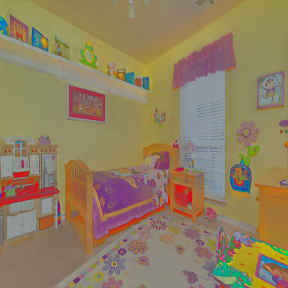}  & 
        \includegraphics[width=0.11\textwidth,height=0.07\textwidth]{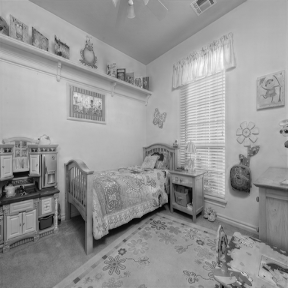}
         \vspace{-1mm}
        \\                       
        {\scriptsize Image} & {\scriptsize Zhou~\cite{zhou2015learning} ($A$)}  & {\scriptsize Zhou \cite{zhou2015learning} ($S$)}  & {\scriptsize CGI \cite{cgintrinsic} ($A$)} & {\scriptsize CGI \cite{cgintrinsic} ($S$)}   & {\scriptsize Fan \cite{fan2018revisiting} ($A$)} & {\scriptsize Fan \cite{fan2018revisiting} ($S$)} & {\scriptsize FFI-Net ($A$)} & {\scriptsize FFI-Net ($S$)}
    \end{tabular}  
  \caption{ Qualitative comparisons of (A)lbedo and (S)hading on IIW. Results of our FFI-Net are visually very similar to the best performing results~\cite{fan2018revisiting}. 
    } \label{fig:iiw} 
\end{figure*}

%% file: fig-wospecloss.tex
\begin{figure}
\begin{center}





\includegraphics[width=\linewidth]{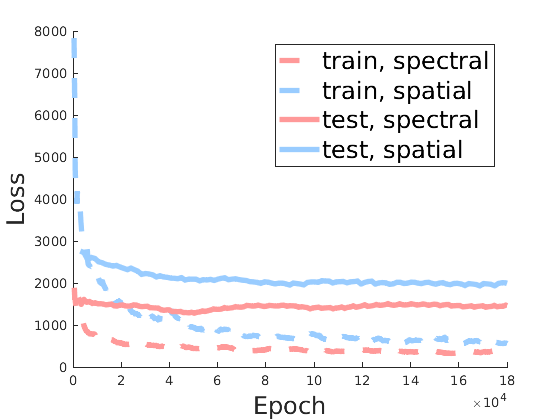}
\end{center}
\caption{
Network training convergence for the proposed spectral (red) and the conventional (blue) residual blocks. 
}
\label{figure:without_spectralloss}
\end{figure}

%% file: conclusion.tex
\section{Conclusion}
We proposed a Fast Fourier Intrinsic Network architecture (FFI-Net)
for intrinsic image decomposition. FFI-Net differs from the existing
works in the sense that inference and network training is
performed in the spectral domain and multi-scale learning is
implemented by spectral banding. The spectral residual blocks
introduced in this work converge faster and provide better accuracy than the conventional residual blocks. FFI-Net is lightweight and does not
need auxiliary networks for training. Our experiments with
three popular datasets verify that FFI-Net achieves state-of-the-art accuracy without auxiliary data training or
specific parameter tuning.
In the future work, we will explore the potential of spectral
network components in other image-to-image translation tasks such as depth estimation and spatial illumination estimation.

\medskip 

\para{Acknowledgement: } This work was partially supported by the National Natural Science
Foundation of China (NSFC) under Grant No. 61828602.